%% file: iclr2027_conference.tex
\documentclass{article} 
\usepackage{iclr2027_conference,times}

\setcitestyle{numbers,square,comma}

\input{preamble}
\title{{\centering\fontfamily{ptm}\selectfont
  {\fontsize{19}{23}\scshape\vspace{0.5em}What Hallucinations Have We Measured?}\\[1pt]
  {\hspace{0.5em}\large\centering Semantic Illusion, Confabulation, and Expressed Uncertainty}}}
\hypersetup{pdftitle={What Hallucinations Have We Measured? Semantic Illusion, Confabulation, and Expressed Uncertainty in LLMs}}

\newcommand{\contributionbanner}{%
  \begin{tcolorbox}[center,colback=takecol!5,colframe=takecol,boxrule=0.4pt,arc=1pt,
    left=5pt,right=5pt,top=3pt,bottom=3pt]
  \small\textbf{Call for contributions.} We believe clarifying what hallucination benchmarks measure is a community effort. If you develop hallucination detection or mitigation methods and would like them evaluated under our taxonomy, or want to extend it to new settings, please get in touch: \href{mailto:hazel.kim@cs.ox.ac.uk}{hazel.kim@cs.ox.ac.uk}.
\end{tcolorbox}}

\author{\begin{minipage}{\dimexpr\textwidth-2\tabcolsep\relax}\centering\normalsize\mdseries
\vspace{1.5em}\textbf{Hazel H. Kim$^{1,*}$\\[10pt] Jakub Binkowski$^{2}$\quad Ekaterina Fadeeva$^{3}$\quad Juan David Guerra$^{4,5}$\quad
Yijie Hao$^{6}$\\[1pt] Zheng Yi Ho$^{7}$\quad Baixiang Huang$^{6}$\quad Buyun Liang$^{8}$\quad Siyuan Liang$^{7}$\quad
Anh Tuan Luu$^{7,9}$\\[1pt] Potsawee Manakul$^{10}$\quad Shahrad Mohammadzadeh$^{11}$ \quad Niels M\"undler-Sasahara$^{3}$\\[1pt] Aidar Myrzakhan$^{12}$\quad Artem Shelmanov$^{12}$\quad Zhiqiang Shen$^{12}$ \quad Kai Shu$^{6}$\quad Ponhvoan Srey$^{7}$\\[1pt]
Dacheng Tao$^{7}$\quad Ren\'e Vidal$^{8}$\quad Haofei Yu$^{13}$\quad Lei Yu$^{14,15}$\quad Hongxiang Zhang$^{16}$\\[10pt]
Yarin Gal$^{1}$\quad Hinrich Sch\"utze$^{17,18}$\quad Philip H.S. Torr$^{1}$}\\[12pt]
$^{1}$University of Oxford \quad $^{2}$Wroclaw University of Science and Technology \quad $^{3}$ETH Zurich\\[1pt]
$^{4}$Mila -- Quebec AI Institute \quad $^{5}$Polytechnique Montr\'eal \quad $^{6}$Emory University\\[1pt]
$^{7}$Nanyang Technological University \quad $^{8}$University of Pennsylvania \quad $^{9}$VinUniversity\\[1pt]
$^{10}$Independent Researcher \quad $^{11}$McGill University \quad $^{12}$MBZUAI \\[1pt] $^{13}$UIUC \quad
$^{14}$University of Toronto \quad $^{15}$Meta (FAIR) \quad $^{16}$Purdue University \\[1pt] $^{17}$LMU Munich \quad $^{18}$Munich Center for Machine Learning\\[6pt]
$^{*}$Correspondence to \texttt{hazel.kim@cs.ox.ac.uk}
\vspace{-1em}
\end{minipage}}
\date{}

\begin{document}

\maketitle
\setcounter{footnote}{0}
\renewcommand{\thefootnote}{\arabic{footnote}}

\begin{abstract}
Hallucination remains one of the most critical challenges in deploying large language models (LLMs) reliably and safely. 
Yet the term is used inconsistently across the research community.
We argue for a strict definition of \emph{hallucinations} in LLMs.
First, hallucinations should be \emph{confident incorrectness}, kept separate from \emph{expressed uncertainty} (\UNK)\footnote{Model responses that explicitly communicate epistemic uncertainty, including both abstention (i.e., declining to provide an answer) and low-confidence answers (e.g., "I am not sure but maybe ...", "I don't know but ..."). }.
\UNK{} reveals that the model may be wrong, whereas a hallucination does not reveal it and is therefore more dangerous in practice. Conflating the two obscures their distinct behaviors and makes them hard to study.
Second, hallucinations should be split by a simple re-elicitation test: does any semantically equivalent prompt recover the correct response?
This test probes whether the relevant knowledge\footnote{We use \textit{knowledge} loosely to mean relevant information stored in the model's latent representations (i.e., pretrained, parametric knowledge). Whether a model truly possesses a piece of knowledge cannot be verified from the outside: it may be stored but unretrievable, or absent from training. We therefore operationally treat knowledge as accessible if some semantically equivalent prompt elicits it without model parameter changes.} is accessible. 
If none does, the knowledge is likely absent or inaccessible, and the error type is a \emph{confabulation} (\CF) that likely requires external knowledge or retraining; if some does, the knowledge is accessible from the model but applied inconsistently, and the error type is a \emph{semantic illusion} (\SI) that simple prompt or decoding changes can mitigate\footnote{Any intervention that revises an originally incorrect model response into a correct one.}.
Re-evaluating 14 recent hallucination methods under this taxonomy yields three findings.
(1)~Mitigation methods are far more effective for \SI{} than for \CF{}.
(2)~Detection methods flag \UNK{} more often than hallucinated responses (i.e., \SI{} and \CF{}).
(3)~Latent signals are more informative than output-level signals for \CF{}. 
An audit of recent hallucination papers shows that most conflate these categories. 
We recommend reporting \CF{}, \SI{}, and \UNK{} separately, and introduce a practical labeling protocol that needs only a modest number of semantically equivalent prompts.
\end{abstract}

\etocdepthtag.toc{main}

\input{sections/introduction}
\input{sections/survey}

\input{sections/taxonomy}

\input{sections/method}
\input{sections/experiment}
\input{sections/conclusion}

\bibliography{iclr2027_conference}
\bibliographystyle{iclr2027_conference}

\etocdepthtag.toc{main}

\newpage
\appendix
\etocdepthtag.toc{appendix}

\begingroup
  \hypersetup{linkcolor=black}
  \etocsettagdepth{main}{none}
  \etocsettagdepth{appendix}{subsection}

  \etocsettocstyle
    {\section*{Appendix Contents}%
     \noindent\rule{\linewidth}{0.4pt}\par\vspace{1.2ex}}
    {\par\vspace{0.6ex}\noindent\rule{\linewidth}{0.4pt}\par}

  \etocsetstyle{section}
    {}
    {\addvspace{1.0ex}}
    {\noindent\makebox[2em][l]{\textbf{\etocnumber}}%
     \textbf{\etocname}\hfill\textbf{\etocpage}\par}
    {}

  \etocsetstyle{subsection}
    {}
    {\addvspace{0.2ex}}
    {\noindent\hspace{2em}\makebox[2.8em][l]{\etocnumber}%
     \etocname\leaders\hbox to 0.8em{\hss.\hss}\hfill\etocpage\par}
    {}

  \etoctoclines  
  \tableofcontents
\endgroup

\input{sections/appendix}

\end{document}

%% file: preamble.tex
\usepackage[utf8]{inputenc}

\usepackage{newunicodechar}
\usepackage{textcomp}
\usepackage{wrapfig}
\usepackage{caption} 
\usepackage{listings}
\usepackage{tikz}

\newcommand{\ecoparagraph}[1]{\par\noindent\textbf{#1}\;}

\usepackage[T1]{fontenc}    
\usepackage{hyperref}       
\usepackage{url}            
\usepackage{booktabs}       
\usepackage{amsfonts}       
\usepackage{nicefrac}       
\usepackage{microtype}      
\usepackage{xcolor}        
\usepackage{amsmath}
\usepackage{threeparttable}
\usepackage{array} 
\usepackage{graphicx}
\usepackage{multirow}
\usepackage{rotating}   
\usepackage{amssymb}    
\usepackage{colortbl}
\usepackage{pdflscape}  
\usepackage{makecell}    
\usepackage{adjustbox}   
\usepackage{longtable}
\usepackage{enumitem}
\usepackage[most]{tcolorbox}
\usepackage{pifont}
\usepackage{subcaption}
\usepackage{placeins} 
\usepackage{tabularx}
\usepackage{array}
\usepackage{xcolor}
\usepackage{makecell}
\usepackage{hyperref}
\usepackage{wrapfig}
\usepackage{caption}
\usepackage{siunitx}
\usepackage[dvipsnames]{xcolor}
\usepackage{caption}
\usepackage[T1]{fontenc}
\usepackage{amsmath, amssymb, amsthm}
\usepackage{thmtools}          
\usepackage[most]{tcolorbox}   
\usepackage{xcolor}
\usepackage{etoc}
\usepackage{booktabs,longtable,array}

\definecolor{headcf}{RGB}{140,40,15}  
\definecolor{headsi}{RGB}{15,70,140} 

\hypersetup{
    colorlinks=true,
    citecolor=headsi,
    linkcolor=headcf,
    urlcolor=headcf
}

\declaretheorem[style=remark, name=Remark]{remark2}

\renewcommand{\thefootnote}{\fnsymbol{footnote}}

\definecolor{checklistbox-color}{HTML}{4F46E5}      
\definecolor{checklistbox-color-bg}{HTML}{EEF2FF}   

\newtcolorbox{checklistbox}[2][]{%
  enhanced,
  breakable,
  sharp corners,
  boxrule=0pt,
  colback=checklistbox-color-bg,
  colframe=checklistbox-color-bg,
  borderline west={3pt}{0pt}{checklistbox-color},
  coltitle=checklistbox-color,
  fonttitle=\bfseries\small,
  title={#2},
  left=8pt, right=6pt, top=4pt, bottom=4pt,
  #1
}

\definecolor{cfcol}{HTML}{D1553F}   
\definecolor{sicol}{HTML}{3F6FB5}   
\definecolor{unkcol}{HTML}{7A7A7A}  
\definecolor{nhcol}{HTML}{4E9A6B}   
\definecolor{takecol}{HTML}{2F4858}

\newcommand{\CF}{{\color{cfcol}\textsc{\textbf{cf}}}}
\renewcommand{\SI}{{\color{sicol}\textsc{\textbf{si}}}}
\newcommand{\UNK}{{\color{unkcol}\textsc{\textbf{unk}}}}
\newcommand{\NH}{{\color{nhcol}\textsc{\textbf{nh}}}}

\newcommand{\VQ}{\mathcal{V}(Q)}
\newcommand{\correct}{\mathrm{correct}}

\newtcolorbox{takeaway}{enhanced,breakable,colback=takecol!5,colframe=takecol,
  boxrule=0pt,leftrule=3pt,arc=0pt,outer arc=0pt,left=7pt,right=7pt,top=4pt,bottom=4pt,
  before skip=8pt,after skip=8pt,
  fontupper=\small,
  overlay unbroken and first={\node[anchor=north west,fill=takecol,text=white,font=\scriptsize\bfseries\sffamily,
    inner xsep=4pt,inner ysep=2pt] at ([xshift=3pt,yshift=0pt]frame.north west) {TAKEAWAY};},
  top=11pt}

\newtcolorbox{defbox}[1]{enhanced,colback=gray!4,colframe=gray!55,boxrule=0.4pt,arc=1pt,
  left=5pt,right=5pt,top=3pt,bottom=3pt,fonttitle=\bfseries\small,coltitle=black,
  colbacktitle=gray!14,title={#1}}

\usepackage{xcolor}
\usepackage{tikz}
\usetikzlibrary{arrows.meta}
\usepackage{pgfplots}
\pgfplotsset{compat=1.17}

\definecolor{cfcol}{HTML}{D1553F}   
\definecolor{sicol}{HTML}{3F6FB5}   
\definecolor{unkcol}{HTML}{7A7A7A}  
\definecolor{nhcol}{HTML}{4E9A6B}   

\theoremstyle{remark}
\newtheorem{remark}{Remark}

\newcommand{\uncertain}{\mathrm{unc}}

\usepackage[most]{tcolorbox}   
\colorlet{hypcol}{-takecol}
\definecolor{hyptag}{HTML}{AD8469}

\newtcolorbox{hypothesis}[2]{enhanced,breakable,
  colback=hypcol!13,colframe=hyptag,
  boxrule=0pt,leftrule=3pt,arc=0pt,outer arc=0pt,
  left=7pt,right=7pt,top=11pt,bottom=4pt,
  before skip=8pt,after skip=8pt,
  fontupper=\small,
  overlay unbroken and first={%
    \node[anchor=north west,fill=hyptag,text=white,
          font=\scriptsize\bfseries\sffamily,inner xsep=4pt,inner ysep=2pt]
      at ([xshift=3pt]frame.north west) {#1\enspace#2};
  }}

%% file: sections/introduction.tex
\section{Introduction}

\begin{figure*}[t]
\centering
\pgfmathsetlengthmacro{\nodew}{30.5mm}
\pgfmathsetlengthmacro{\colsep}{(\textwidth-\nodew-0.6pt)/3}
\pgfmathsetlengthmacro{\xa}{\nodew/2}
\pgfmathsetlengthmacro{\xb}{\xa+\colsep}
\pgfmathsetlengthmacro{\xc}{\xb+\colsep}
\pgfmathsetlengthmacro{\xd}{\xc+\colsep}
\begin{tikzpicture}[
  font=\sffamily\footnotesize,
  box/.style={rectangle,rounded corners=1.5pt,align=center,inner xsep=3pt,inner ysep=4pt},
  q/.style={box,font=\sffamily\scriptsize,text width=21.5mm,inner xsep=0pt,minimum width=26mm,
            inner ysep=5pt,minimum height=11mm,draw=black!35,fill=black!2,line width=0.4pt},
  o/.style={box,text width=\nodew-6pt-0.6pt,minimum height=14mm,draw=#1,fill=#1!7,line width=0.6pt},
  hal/.style={line width=1.4pt,text width=\nodew-6pt-1.4pt},
  lab/.style={font=\sffamily\scriptsize,text=black!55,inner sep=0pt},
  arr/.style={-{Stealth[length=4pt,width=3.2pt]},black!50,line width=0.5pt,
              shorten <=1.5pt,shorten >=1.5pt}]

  \node[q] (q0) at (\xa,0) {Original prompt\\[1pt]answered correctly?};
  \node[q] (q1) at (\xb,0) {Any re-elicitation\\[1pt]answered correctly?};
  \node[q] (q2) at (\xc,0) {Expresses uncertainty under\\[1pt]any condition?};

  \node[o=nhcol]      (nh)  at (\xa,-2.4) {\textbf{\color{nhcol}No hallucination}\\[2pt]{\scriptsize\color{black!60}correct under the\\original prompt}};
  \node[o=sicol,hal]  (si)  at (\xb,-2.4) {\textbf{\color{sicol}Semantic illusion}\\[2pt]{\scriptsize\color{black!60}hallucination,\\recoverable}};
  \node[o=unkcol]     (unk) at (\xc,-2.4) {\textbf{\color{unkcol}Expressed uncertainty}\\[2pt]{\scriptsize\color{black!60}admits it is unsure;\\not a hallucination}};
  \node[o=cfcol,hal]  (cf)  at (\xd,-2.4) {\textbf{\color{cfcol}Confabulation}\\[2pt]{\scriptsize\color{black!60}hallucination,\\unrecoverable}};

  \foreach \a/\b in {q0/nh,q1/si,q2/unk}
    \draw[arr] (\a.south) -- node[lab,right=4pt]{yes} (\b.north);
  \draw[arr] (q0.east) -- node[lab,above=3pt]{no} (q1.west);
  \draw[arr] (q1.east) -- node[lab,above=3pt]{no} (q2.west);
  \draw[arr] (q2.east) -| node[lab,above=3pt,pos=0.3]{no} (cf.north);

  \node[font=\sffamily\footnotesize,text=black!65,anchor=north west,inner sep=0pt]
        at (0,-3.55) {Re-elicitation: paraphrase, instruction or decoding change (without additional information)};
\end{tikzpicture}
\caption{\textbf{Proposed taxonomy of model responses.} An incorrect answer is a semantic illusion if any semantically equivalent re-elicitation recovers the correct answer. If none does, it is a confabulation, unless the model expresses uncertainty under some condition, in which case it is \UNK{}. Only \SI{} and \CF{} are hallucinations.}
\label{fig:overview}
\vspace{-1em}
\end{figure*}

Hallucinations in large language models (LLMs) have emerged as a central concern for both researchers and practitioners. Yet there is surprisingly little consensus on what the term actually means.
Across the literature, it is applied to a broad and heterogeneous set of model failures---factual inaccuracies~\cite{rawte2023survey, Dang25survey, Alansari26survey},  faithfulness errors~\cite{Ji23survey, zhang-etal-2025-sirens}, reasoning inconsistencies~\cite{Huang25survey, Chakraborty25survey}, and uncertain outputs~\cite{zhang-etal-2025-sirens, Huang25survey, xia-etal-2025-survey}---often without distinction or explicit definition. This conceptual ambiguity is not merely terminological but obscures the underlying causes of model incorrect responses, undermines evaluation validity, and hinders the development of targeted mitigation strategies.

We propose one definition and one test (Fig.~\ref{fig:overview}). A hallucination is a confident incorrect response, so a model that says ``I don't know'' is expressing uncertainty (\UNK), not hallucinating. Hallucinations then split by whether the knowledge is \emph{behaviorally accessible}. If some paraphrase, instruction or decoding setting recovers the correct answer, the error is a semantic illusion (\SI); if none does, it is a confabulation (\CF) (see \S\ref{sec:definition} for terminological motivations). The test is an operational proxy for knowledge accessibility, and it makes a sharp prediction that \SI{} should yield to inexpensive inference-time mitigation, while \CF{} should need new, additional knowledge.

Our main contributions and findings are as follows:
\begin{itemize}
  \item \textbf{An audit of the community} (\S\ref{sec:community_survey}). Only 35\% of recent hallucination papers define hallucination as confident incorrectness. Yet 95.8\% of surveyed first authors say it should be distinguished from uncertainty, and only 16.7\% report that they did so.
  \item \textbf{A behavioral taxonomy} (\S\ref{sec:definition} and \S\ref{sec:taxonomy}) We define hallucination as a confidently incorrect response, separate from expressed uncertainty (\UNK), and introduce two error types based on a re-elicitation test: \emph{semantic illusions} (\SI), which some semantically equivalent re-elicitation corrects, and \emph{confabulations} (\CF), which none does. We use this taxonomy to re-evaluate 14 hallucination methods on two LLMs.
  \item \textbf{Confabulations are a blind spot} (\S\ref{sec:f1}). Hallucination methods struggle to detect \CF{}, especially under uncertainty elicitation, and \CF{} is the hardest error type to mitigate, even with external knowledge.
  \item \textbf{Hallucination methods mostly measure expressed uncertainty} (\S\ref{sec:f2}). They score \UNK{}, which is not a hallucination, higher than \SI{} and \CF{}.
  \item \textbf{The right uncertainty representations depend on the error type} (\S\ref{sec:f3}). Verbal signals suit \UNK{}, semantic ones \SI{}, and latent ones \CF{}.
\end{itemize}

\newcommand{\findrow}[4]{%
  \begin{tcolorbox}[enhanced,blanker,left=0pt,right=0pt,top=1pt,bottom=1pt,
     sidebyside,sidebyside align=center seam,lefthand width=0.72\linewidth,sidebyside gap=8pt]
  \begin{tabular}{@{}p{16pt}@{\hspace{6pt}}p{0.64\linewidth}@{}}
  \raisebox{-0.5\height}{\tikz\node[circle,fill=takecol,text=white,font=\sffamily\bfseries\small,
     minimum size=15pt,inner sep=0pt]{#1};} &
  \raggedright\small\textbf{#2}\par\footnotesize #3
  \end{tabular}
  \tcblower
  \centering #4
\end{tcolorbox}}

These effects are invisible in an aggregate hallucination score. An audit of recent papers (\S\ref{sec:community_survey} and \S\ref{app:survey}) shows such aggregation is the norm. We also contribute a labeling protocol (\S\ref{sec:taxonomy}) that stabilizes at about ten paraphrases, so the taxonomy is plausible to adopt.

%% file: sections/survey.tex
\section{What the Community Measures}
\label{sec:community_survey}

\begin{figure}[t!]
    \centering
    \vspace{-1em}
    \includegraphics[width=0.95\linewidth]{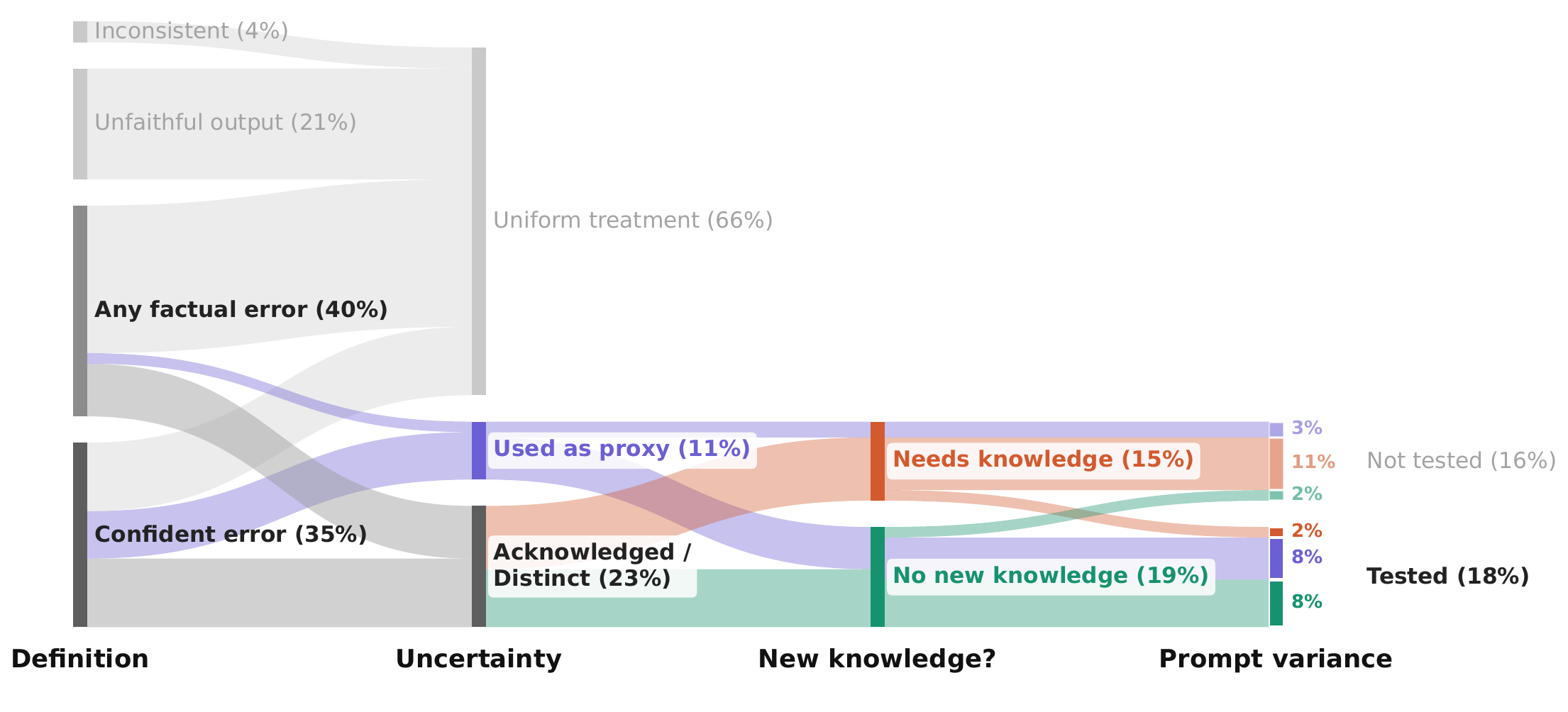}
    \vspace{-1em}
    \caption{\textbf{Recent text-based hallucination papers filtered by their definitions and approaches with uncertainty estimation, additional knowledge, and prompt variance.} From a pool of all qualifying text-based papers$^{*}$ across ICML, NeurIPS, ICLR, ACL, and EMNLP (2021--2025), most papers neither distinguish hallucination from uncertainty nor test prompt variance before considering external knowledge as a source. The methods that skip these distinctions risk targeting the wrong error type entirely. The detailed explanations are in \S\ref{sec:community}.}
    \label{fig:hallucination_funnels_main}
    \vspace{3pt}
    {\hspace{1em}\footnotesize\raggedright$^{*}$We did not filter papers further if they use the term ``hallucination'' in the title and take a text-based approach.\par}
    \vspace{-1em}
\end{figure}

We audit all the text-based papers with ``hallucination'' in the title from five major venues (2021--2025; \S\ref{sec:community}) and survey 24 of their first authors (\S\ref{app:survey}).
The paper categorization is illustrated in Fig.~\ref{fig:hallucination_funnels_main}. Two gaps stand out.

\paragraph{Hallucination is rarely separated from uncertainty.}
Only 35\% of papers define hallucination as a confident or plausible error, and only 23\% treat hallucination detection as distinct from uncertainty estimation.
Yet 95.8\% (23/24) of surveyed authors said the two should be distinguished; only 16.7\% (4/24) said their own work did so.

\paragraph{Additional knowledge is added before inexpensive inference-time mitigation methods are tried.}
Of papers concluding that external knowledge is needed (15\% of the 91), most (11\% of the 91) never
tested whether the error disappears under a different phrasing, instruction or decoding setting.
Survey respondents ranked knowledge availability, whether a model lacks a fact or misapplies it, as
the most important dimension of hallucination (79.2\%, 19/24).

\begin{takeaway}
Most hallucination papers measure one undifferentiated error rate. They rarely separate confident incorrectness from expressed uncertainty, and rarely check whether the incorrect response survives rephrasing before adding knowledge, even though many authors think both distinctions matter.
\end{takeaway}

\section{Background}

\ecoparagraph{Definition}
Despite its widespread usage, the term hallucination lacks a consensus definition across the research community. The proportion of papers explicitly defining it has declined steadily since 2021 \cite{narayanan-venkit-etal-2024-audit}. Early NLP usage anchored the term to faithfulness failures, describing generated content unfaithful to a provided source \cite{Ji23survey, zhang-etal-2025-sirens}. With the rise of LLMs, definitions have diverged along two axes. The first concerns the nature of \textbf{\emph{incorrectness}}: from source unfaithfulness \cite{Ji23survey, zhang-etal-2025-sirens}, to real-world unfactualness covering fictional, misleading, or fabricated outputs \cite{rawte2023survey, Dang25survey, Alansari26survey}, and further to outputs conflicting with instructions, context, or logical consistency \cite{Chakraborty25survey, Huang25survey}. The second concerns the degree of \textbf{\emph{plausibility}}: from outputs that simply appear fluent and coherent \cite{Alansari26survey, Dang25survey}, to those whose plausibility is treated as an unavoidable consequence of how LLMs generate text \cite{WanJing25survey, li25survey}, to the strictest view 
requiring misplaced confidence, restricting hallucination to 
assertions the model delivers with unwarranted certainty 
\cite{jiang2024survey, narayanan-venkit-etal-2024-audit, Huang25survey}.

\ecoparagraph{Terminology}\label{background:terminology} The terminological ambiguity has prompted calls for more precise alternatives. A systematic study of researcher perspectives \cite{narayanan-venkit-etal-2024-audit} found that 40.46\% prefer \emph{fabrication}, while others advocate for \emph{confabulation}, a clinical term for false narratives produced to bridge knowledge gaps without intent to deceive \cite{Berlyne_1972, Moscovitch89, Schnider08}. This preference reflects a broader concern that models frequently fabricate responses without sufficient grounds. 
The clinical origin of the term, however, is more specific, denoting errors arising from missing or inaccessible knowledge \cite{Berlyne_1972, Moscovitch89, Schnider08, ChengMWLKD24, marone2023data}. Under this view, responses may vary in form across semantically equivalent prompt variants, yet remain incorrect regardless of how the question is phrased.
This captures only one error type. We therefore retain \emph{hallucination} as the cover term and decompose it into two behaviorally distinct subtypes, \emph{confabulation} and \emph{semantic illusion}, whose empirically separable signatures we define in \S\ref{subsec:distinction} and analyze in \S\ref{sec:results}.

\ecoparagraph{Taxonomy}
Early survey work established the foundational intrinsic/extrinsic taxonomy, where intrinsic hallucinations contradict the source content and extrinsic hallucinations introduce unverifiable information \cite{Ji23survey}. Subsequent work expanded this line of taxonomies. \cite{zhang-etal-2025-sirens} proposed input-conflicting, context-conflicting, and fact-conflicting categories to capture misaligned responses beyond source documents, while \cite{Huang25survey, tonmoy2024survey} separated factuality hallucination from faithfulness hallucination across a hierarchical set of subtypes. 
\cite{Dang25survey} attributed hallucinations to either prompting or model-internal factors by measuring how errors change across prompt variants and across models. 
Unlike prior taxonomies based on triggers or contradictions, we classify hallucinations by prompt sensitivity, distinguishing consistent incorrectness (\textit{confabulation}) from prompt-sensitive incorrectness (\textit{semantic illusion}), and linking each to corresponding detection and mitigation strategies.

\section{Hallucination Types}
\label{sec:definition}

We categorize hallucinated responses into two qualitatively different subtypes: \textit{confabulation} and \textit{semantic illusion}.
In this section, we explain where each term comes from and why it fits the corresponding error type.

\ecoparagraph{Type\#1: Confabulation} 
A hallucinated response is a confabulation (\CF{}) if an LLM produces the incorrect output regardless of any inference-time changes. We hypothesize this is due to absent or inaccessible memory relevant to the factual information required by the input.
Confabulation classically refers to a systematic pattern of generating false narratives to bridge knowledge gaps while maintaining high confidence in the fabrications ~\cite{Moscovitch89, Schnider08}. 
It is commonly used as a clinical term for verbal statements that are inconsistent with the patient's history and current situation due to brain damage~\cite{Berlyne_1972, moscovitch1995confabulation, johnson1998false, bartolomeo2017space, french2009false}.
Consequently, confabulation can be associated with the absence of relevant memories in human cognition.
While LLMs operate through fundamentally different mechanisms, they exhibit analogous patterns of generating false information.
One cause is the absence of necessary information in the training data. It is particularly evident when models with specific knowledge cut-off dates lack current information~\cite{ChengMWLKD24, xieWHZGPFD24}. Such a phenomenon is known as \textit{absent memory}. 
Another cause occurs when relevant information exists in the training data but remains underlearned~\cite{marone2023data, pushkarna2022data}.
In these contexts, confabulation manifests as a persistent form of fabrication that resists correction through prompt engineering or inference-time hyperparameter tuning such as random seeds, temperatures, or top-$p$ distributions of the decoding mechanisms.

\ecoparagraph{Type\#2: Semantic Illusion}
A hallucinated response is a semantic illusion (\SI{}) if an LLM produces an incorrect output despite being able to correctly answer the same or semantically equivalent inputs that require the same underlying factual knowledge, without any change to the model parameters (i.e., without further training).
Unlike confabulation, semantic illusion occurs even if the eligible memory has the relevant information.
The terminology is well-illustrated by the classic linguistic example in human cognition where individuals readily answer "two" when asked "how many animals of each kind did Moses take on the ark?", despite knowing that it was Noah who built the ark~\cite{erickson1981words, bredart1988moses}.
The illustration allows us to pinpoint the two major features of semantic illusion:
(1) we choose the erroneous answers even though we know the fact;
(2) we can answer correctly under a different or even exactly the same phrasing.
While the concept of semantic illusion has been studied in human cognition, it has not been formally defined or widely explored in the context of language models. 
In particular, next-token prediction can be dominated by contextual associations that make an answer appear semantically appropriate within the immediate prompt, even when it conflicts with the correct underlying facts.
This is particularly pronounced in chain-of-thought reasoning, where  language models fail to backtrack and reconsider the discrepancy from accurate facts~\citep{zhang2024backtracking, yu2024unveiling}.
It manifests through the model tendency to "snowball" incorrect information, elaborating on flawed premises to maintain the initial semantic coherence~\citep{ZhangPMLS24}.

\section{Problem Formulation}

We formalize the two central hallucination dimensions -- \textit{incorrectness} and \textit{plausibility} -- through the following research questions (RQ) and hypotheses (H).

\newtcolorbox{definitionbox}[1]{
  enhanced,
  colback=black!5,
  frame hidden,
  borderline west={2.5pt}{0pt}{black!50},
  arc=4pt,
  left=10pt,
  right=10pt,
  top=8pt,
  bottom=8pt,
  fonttitle=\bfseries,
  coltitle=black,
  title=#1,
  attach title to upper,
  after title={\hspace{6pt}}, 
  separator sign none,
}

\subsection{RQ1 (Incorrectness): Do hallucinations arise from distinct error types?}
\label{subsec:distinction}
The most basic condition for hallucinations is incorrectness, whether measured against a source~\cite{Ji23survey, zhang-etal-2025-sirens}, the real-world facts~\cite{rawte2023survey, Dang25survey, Alansari26survey}, or an instruction~\cite{Huang25survey, Chakraborty25survey}.
Incorrect outputs, however, are not uniform in how they arise or how they respond to elicitation. 
Accordingly, the difference that matters most in practice is consistency of occurrence.
Yet some incorrect responses can be corrected under a semantically equivalent phrasing (\textit{semantic illusion}), while others recur whatever phrasing is used (\textit{confabulation}).
We ask whether these reflect distinct error types, and whether that distinction changes which detection and mitigation strategies apply.

The difference also invites an epistemic reading, since a response that becomes correct under a semantically equivalent phrasing shows the relevant knowledge is accessible without changing model weights or injecting additional information.
We therefore treat knowledge as behaviorally accessible when some tested condition elicits a correct response. 
The conditions we admit that establish accessibility are semantically equivalent phrasings, instruction variants, decoding configurations (e.g., temperature, nucleus sampling), which require no access to model internals and add no additional hints or context.

\begin{hypothesis}{H1}{Testable Knowledge Accessibility}
We can measure model (accessible) knowledge by checking if any rephrasing,
instruction, or decoding setting makes it answer correctly, and the resulting labels
(\SI{} \&\CF{}) should be stable enough to study.
\end{hypothesis}

\begin{hypothesis}{H2}{Detection and Intervention Asymmetry}
If confabulation (\CF{}) and semantic illusion (\SI{}) are genuinely distinct error types, existing detection and mitigation methods should perform asymmetrically across them. 
\end{hypothesis}

\subsection{RQ2 (Plausibility): Does hallucination require confident incorrectness?}
\label{subsec:rq2}
We define plausibility as confidence, because (i) current LLMs are fluent by default, so surface plausibility is satisfied by nearly every output and draws no useful boundary, and (ii) a confidently stated incorrect response gives the reader no cue to check it, which makes it the more dangerous error.
We say a response is \textbf{confident} if it does not explicitly express uncertainty. 
A confidently incorrect response (\CF{}, \SI{}) and an expressed uncertainty (\UNK{}) may both be scored as incorrect, but reflect qualitatively different behaviors and risks. Uncertainty estimation methods were originally developed to quantify epistemic uncertainty as a signal for detecting incorrect outputs, whereas \UNK{} is a \emph{behavior} in which the model explicitly abstains. As LLMs increasingly exhibit this behavior, the remaining errors are predominantly confident ones, where output-level uncertainty signals have least to detect. We ask whether confidence is a necessary condition for hallucination and whether existing methods respect this distinction in practice.
\begin{hypothesis}{H3}{Hallucination and Uncertainty}
If the hallucination detection methods operates as defined, it should detect confidently incorrect responses.
\end{hypothesis}

\begin{hypothesis}{H4}{Representation Asymmetry} 
Among uncertainty estimation methods,
verbal representations will be most effective at detecting \UNK{}, which carries an explicit uncertainty cue in its output. 
In contrast, \CF{} carries no such cue.
Therefore, requires some intervention beyond the output level.
\end{hypothesis}

%% file: sections/taxonomy.tex
\section{A Behavioral Taxonomy of Incorrect Responses}
\label{sec:taxonomy}

\subsection{Definitions}

Fix a model $M$ and a question $Q$ with gold answer $A$. Let $\VQ$ be a finite set of \emph{test
conditions}: the product of semantically equivalent paraphrases (adding no hints), instruction
variants, and decoding configurations, excluding the original prompt $c_0$. Each response $r(c)$ is
exactly one of: correct, an expressed uncertainty ($\uncertain(r(c))=1$), or a committed incorrect response.

\begin{defbox}{Definition 1 \;(Label partition; question-level and model-specific)}
If $\correct(r(c_0))=1$, the question is \NH. Otherwise it receives exactly one label:
\begin{align*}
\SI:\quad & \exists\, c:\ \correct(r(c))=1;\\
\UNK:\quad & \forall\, c:\ \correct(r(c))=0 \ \text{ and }\ \exists\, c:\ \uncertain(r(c))=1;\\
\CF:\quad & \forall\, c:\ \correct(r(c))=0 \ \text{ and }\ \uncertain(r(c))=0,
\end{align*}
with all quantifiers over $c \in \{c_0\} \cup \mathcal{V}(Q)$. \CF{} and \SI{} are hallucinations; \UNK{} is not.
\end{defbox}

\begin{remark}[Labels are relative to $\VQ$]
\label{rem:relative}
Enlarging $\VQ$ can move a question from \CF{} to \SI{} but never back, because a correct answer
already observed cannot be revoked. The \CF{} rate is therefore an \emph{upper bound} on behaviorally
inaccessible knowledge, and the labels are diagnostics, not intrinsic model properties.
\end{remark}

Formal definitions of each term are in \S\ref{app_sec:definitions}.

\subsection{Is Accessible Knowledge Measurable? Behavioral Evidence from $\mathcal{V}(Q)$}
\label{sec:spectrum}

We define knowledge as behaviorally \emph{accessible} if at least one variant $v \in \mathcal{V}(Q)$ elicits a correct response without additional hints or parameter changes (see \S\ref{app_sec:accessible_knowledge} for details).
To test this, we evaluate each question under its 30 paraphrases, instruction variants, and five decoding settings
(greedy; $T\in\{0.2,0.7\}\times p\in\{0.80,0.95\}$). 
The instruction variants test two different schemes:
\textbf{S1} uses three answer-only instructions; \textbf{S2} adds one that invites uncertainty elicitation (``If uncertain, say: I don't know''). 
Labels are computed per evaluated model.
The labeling schemes are never revealed to  the methods, so detection or mitigation methods are evaluated fairly, seeing only the original questions.
Three observations support the measurability of accessible knowledge.

\begin{figure}[h!]
  \centering
  \vspace{-0.5em}
  \includegraphics[width=0.9\linewidth]{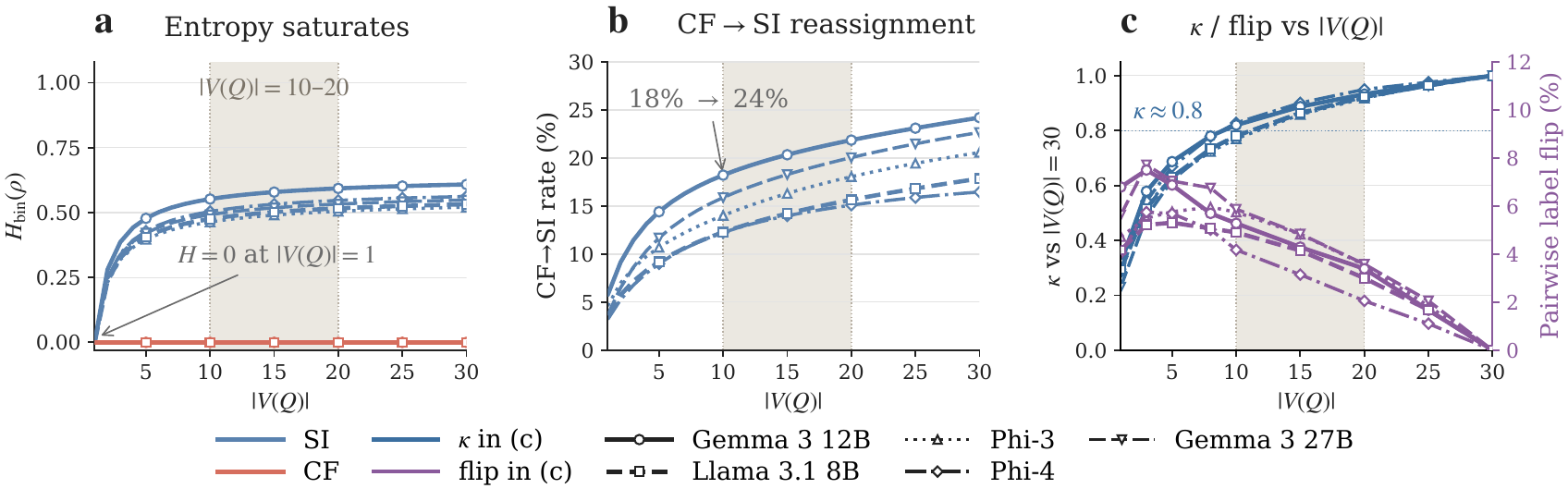}
  \vspace{-0.5em}
  \caption{\textbf{Labels stabilize at about ten paraphrases}, subsampling from a pool of 30.
  (a)~Binary entropy of the per-question correctness rate. (b)~\CF$\to$\SI{} reassignment rate.
  (c)~Cohen's $\kappa$ against the full-pool labels, and pairwise flip rate. Shading marks the
  recommended window. Phi-3, Phi-4 and Gemma~3~27B show that the curves generalize across families.}
  \label{fig:vq}
\end{figure}

\begin{figure}[h!]
    \centering
    \vspace{-.5em}
    \includegraphics[width=\linewidth]{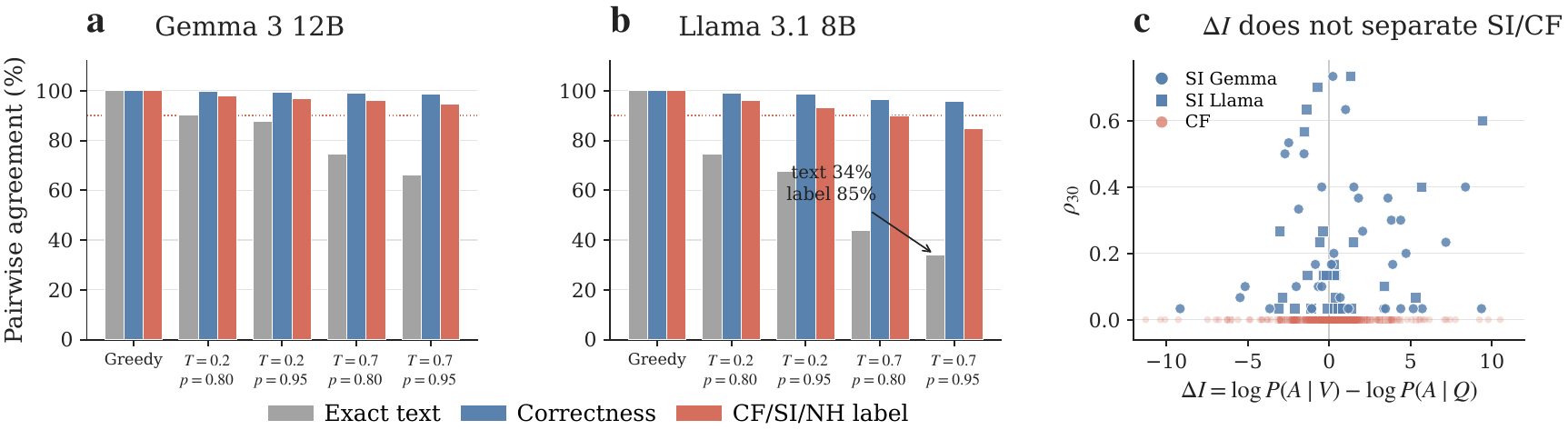}
    \vspace{-1.5em}
    \caption{\textbf{Label stability under decoding, and SI/CF is not an information-injection artifact.} (a)\&(b)~Pairwise agreement (\%) on a 30 question variants, aggregated over three instruction prompts. (c) $\rho_{30}$ vs. the information quantity emerged from paraphrases $\Delta I=\log P(A\mid V)-\log P(A\mid Q)$.} 
    \label{fig:pairwise_agreement_main}
\end{figure}

\ecoparagraph{We can define the sufficiency of $\mathcal{V}(Q)$ to measure the accessible knowledge.}
We observe that the entropy of \SI{} correctness plateaus, \CF$\to$\SI{} reassignment flattens (18\% to 24\% across the $|\VQ|$ (Fig.~\ref{fig:vq}b), and the agreement with full-pool labels reaches $\kappa\approx0.8$ around $|\VQ|=10$ (Fig.~\ref{fig:vq}c).
The stability of the measured knowledge accessibility directly impacts labeling and we evaluated the label agreement across different decoding schemes. 
At $T{=}0.7$, $p{=}0.95$, both gemma and llama models show stable label agremeent. Especially, llama models do not provide the exactly the same responses in terms of textual forms (across conditions only 34\% of the time), but its labels agree 85\% of the time  (Fig.~\ref{fig:pairwise_agreement_main}b).

\ecoparagraph{Correct responses under semantically equivalent prompts reflects accessibility, not added information.}
The information a semantically equivalent question adds about the label information, $\Delta I=\log P(A\mid V)-\log P(A\mid Q)$, is centered near zero for both \SI{} and \CF{} questions, yet only \SI{} questions are ever answered correctly (Fig.~\ref{fig:pairwise_agreement_main}c; median $\rho_{30}=0.17$ / $0.13$ for Gemma / Llama). This shows that both \SI{} and \CF{} have almost no additional information or context on top of their original question and has an equal, trivial amount of additional information if any.

\begin{takeaway}
\textbf{H1 is supported.} Accessible knowledge can be measured from behavior alone: \SI{}
questions are answered correctly under some rephrasings that add no information
($\Delta I \approx 0$), while \CF{} questions never are. The resulting labels are stable
enough to study, converging at about ten paraphrases ($\kappa \approx 0.8$) and agreeing
across decoding settings even when the response text does not.
\end{takeaway}

\subsection{Labeling protocol and validation}
\label{subsec:labeling_protocol}

\ecoparagraph{Recommended labeling mechanism.}
(1)~Use approximately 10 variants ($|\mathcal{V}(Q)| \approx 10$) as a default and label under greedy or low-temperature decoding.
Because \textsc{si} and \textsc{cf} labels remain largely stable across decoding settings even when the exact response text changes (Fig.~\ref{fig:pairwise_agreement_main}a,b), we can take this setup as a practical approximation of the full labeling conditions.
(2)~Report the $H_{\mathrm{bin}}$ (and $\kappa$ / flip-rate) convergence curve on one's own data as a sufficiency check.
(3)~Treat \textsc{cf} as an upper-bound on behavioral inaccessibility.
Additional variants can only move \textsc{cf} error cases to \textsc{si} ones.
These steps makes the taxonomy reproducible across studies with a modest number of standard paraphrases.

\begin{takeaway}
Labeling is plausible: about ten paraphrases under greedy decoding give stable labels. Report the convergence curve as a sufficiency check, and read the \CF{} rate as an upper bound on behavioral inaccessibility.
\end{takeaway}

%% file: sections/method.tex
\section{Experimental Setup}
\label{sec:setup}

\paragraph{Study design.} We evaluate 14 detection and mitigation methods from 2021--2025 (Table~\ref{tab:methods}; See \S\ref{sec:methods} for detailed explanation and \S\ref{app_dimensions_notation} for full design dimensions). 
For RQ1, we examine prompt sensitivity (whether prompt variation is modeled), intervention stage (inference-time vs. post-hoc vs. external knowledge), and model knowledge assumption (whether the method implicitly assumes knowledge is absent or present). For RQ2, we examine the usage of uncertainty estimation (whether uncertainty is used for calibration or correction) and uncertainty representation (verbal, latent, semantic, or none). We have verified faithful implementations by asking the authors of those papers to conduct the experiments.

\begin{table}[h!]
\centering
\caption{\textbf{The 14 methods, grouped by the two dimensions used in the main text.} Other
groupings are analyzed in \S\ref{sec:methods}.}
\label{tab:methods}
\small
\begin{tabularx}{\linewidth}{@{}l X@{}}
\toprule
\multicolumn{2}{@{}l}{\emph{Intervention stage} (Finding 1)}\\
Inference-time & ActLCD \citep{zhang-etal-2025-active}, NoVo \citep{ho2025novo} \\
Post-hoc & SelfCheckGPT \citep{manakul-etal-2023-selfcheckgpt}, UQ Heads \citep{shelmanov-etal-2025-head} \\
External knowledge & DRAG \citep{chen-etal-2025-drag}, HEB \citep{huang2025can}, CS \citep{hao-etal-2025-beyond} \\
\midrule
\multicolumn{2}{@{}l}{\emph{Uncertainty representation} (Findings 2--3)}\\
Verbal & VUF \citep{ji-etal-2025-calibrating}, UQ Heads \citep{shelmanov-etal-2025-head} \\
Latent & UQ Heads \citep{shelmanov-etal-2025-head}, SenD \citep{mohammadzadeh-etal-2025-hallucination}, IRIS \citep{srey-etal-2025-unsupervised}, Attn.\ Sink \citep{binkowski2026attentionsinksinternalsignals} \\
Semantic & SelfCheckGPT \citep{manakul-etal-2023-selfcheckgpt}, SCH \citep{mundler2024selfcontradictory}, IRIS \citep{srey-etal-2025-unsupervised} \\
None & SECA \citep{liang2025seca}, DRAG \citep{chen-etal-2025-drag}, HEB \citep{huang2025can}, NoVo \citep{ho2025novo}, ActLCD \citep{zhang-etal-2025-active}, LapEigvals \citep{binkowski-etal-2025-hallucination} \\
\bottomrule
\end{tabularx}
\end{table}

\paragraph{Data and models.} Closed-book factual QA from TriviaQA \citep{joshi2017triviaqa} and
SimpleQA \citep{wei2024measuring}, evaluated on Llama~3.1~8B and Gemma~3~12B.
Closed-book QA gives verifiable answers and makes parametric knowledge the only source of the answer. See \S\ref{app:dataset} for detailed explanations on dataset construction pipeline, including prompt templates.

\begin{wraptable}{r}{0.4\textwidth}
\centering
\caption{\textbf{Label distribution}
($|\VQ|=10$). Uncertainty elicitation (S2) converts most \CF{} into \UNK.}
\vspace{-0.5em}
\label{tab:labels}
\footnotesize
\begin{tabular}{@{}lccc@{}}
\toprule
Scheme & \CF & \SI & \UNK \\
\midrule
S1 & 0.36 & 0.62 & 0.02 \\
S2 & 0.04 & 0.64 & 0.32 \\
\bottomrule
\end{tabular}
\vspace{-1em}
\end{wraptable}

\paragraph{Metrics.} With the method's hallucination prediction $h$, correctness $c$ and the label $\ell$, we report \emph{recall} $P(h{=}1\mid c{=}0,\ell)$ and \emph{residual error rate} $P(c{=}0 \mid h{=}0, \ell)$, for detection.  For mitigation, we report \emph{mitigation correctness} $P(c{=}1\mid h{=}1,\ell)$, the proportion of detected errors that are corrected. Results are averaged over methods within a group. 
We also report the mean of hallucination scores $r \in [0,1]$ when the method provides them.

%% file: sections/experiment.tex
\section{Finding 1: Confabulations Are a Blind Spot for Detection and Mitigation}
\label{sec:f1}

\begin{figure}[h!]
    \centering
    \vspace{-1em}
    \includegraphics[width=0.7\linewidth]{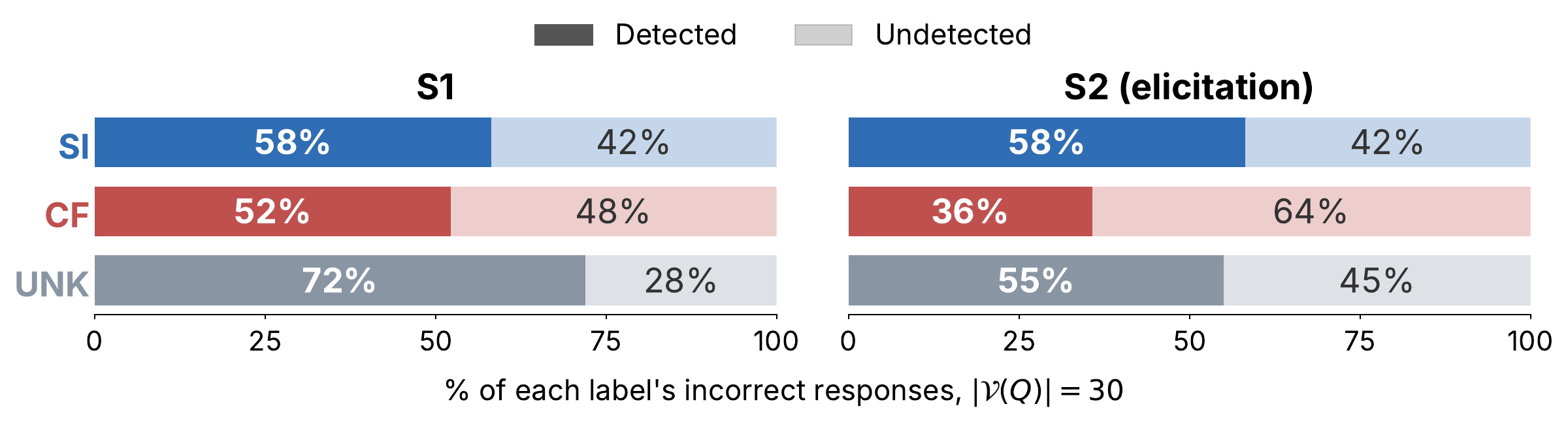}
    \vspace{-0.5em}
    \caption{\textbf{Proportion of incorrect responses in each label detected (dark) and undetected (light)} at $|\mathcal{V}(Q)|=30$, without (S1) and with (S2) uncertainty elicitation across all methods.}
  \label{fig:h2_detection_bars}
  \vspace{-0.5em}
\end{figure}

\ecoparagraph{The detection gap.} Fig.~\ref{fig:h2_detection_bars} shows the detection asymmetry among different labels. Without uncertainty elicitation (S1), \CF{} responses are detected 6\% less than \SI{} responses. With elicitation (S2), the gap widens. Elicitation barely changes \SI{} detection, but lowers detection of both \CF{} and \UNK{}.

\begin{figure}[h!]
    \centering
    \includegraphics[width=0.6\linewidth]{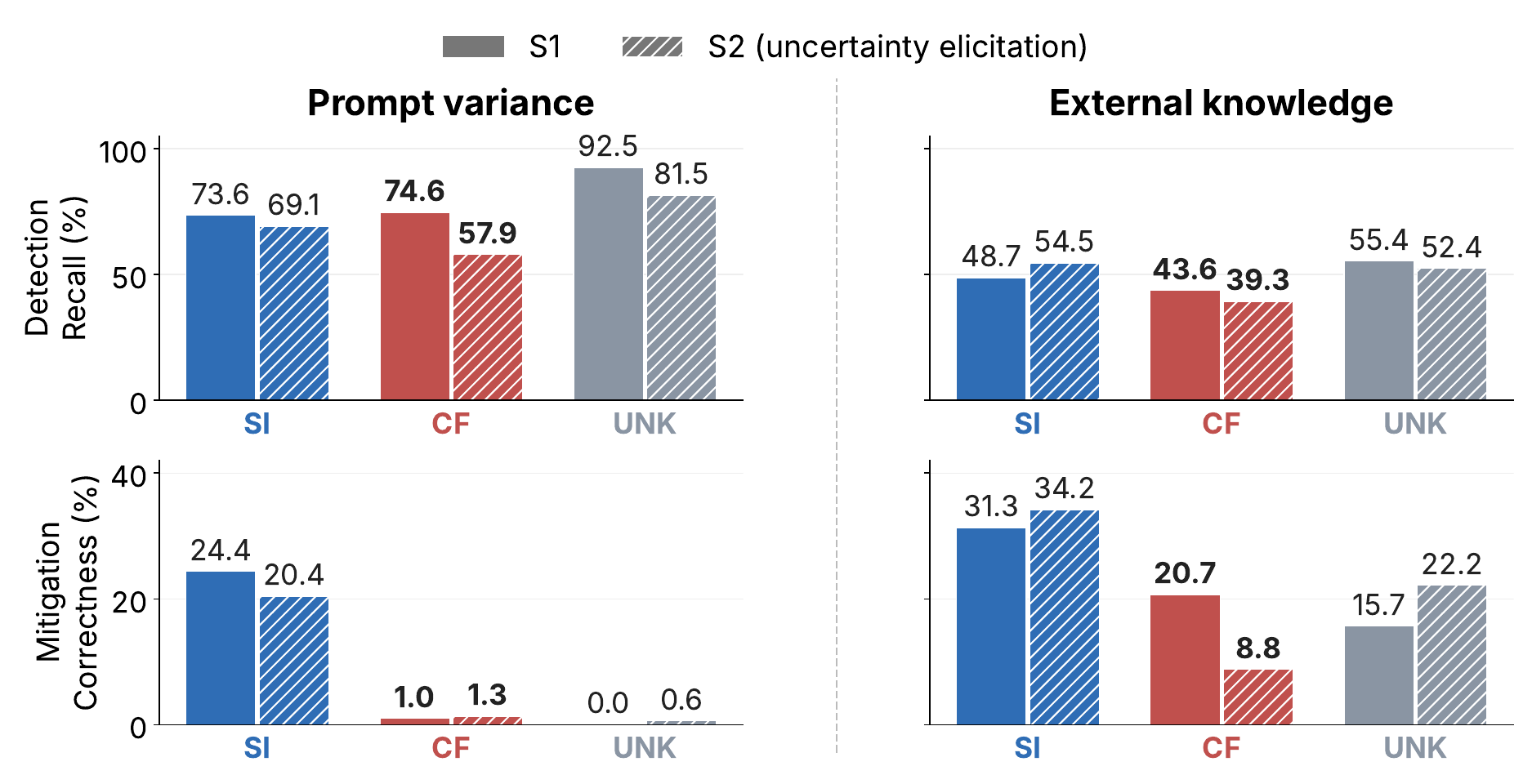}\\
    \vspace{-0.5em}
    \caption{\textbf{\CF{} blind spot} both for methods targeting prompt variance and external knowledge}
  \label{fig:h2_bars}
\end{figure}

\ecoparagraph{The mitigation gap.}
What is interesting is that \CF{} is hard to mitigate not only for prompt-variance methods, as expected, but also for external-knowledge methods, which should in principle provide the absent knowledge, despite their larger overall mitigation effect (Fig.~\ref{fig:h2_bars}). This suggests that what matters is whether the missing knowledge is actually provided.

\begin{figure}[h!]
  \centering
    \includegraphics[width=0.5\linewidth]{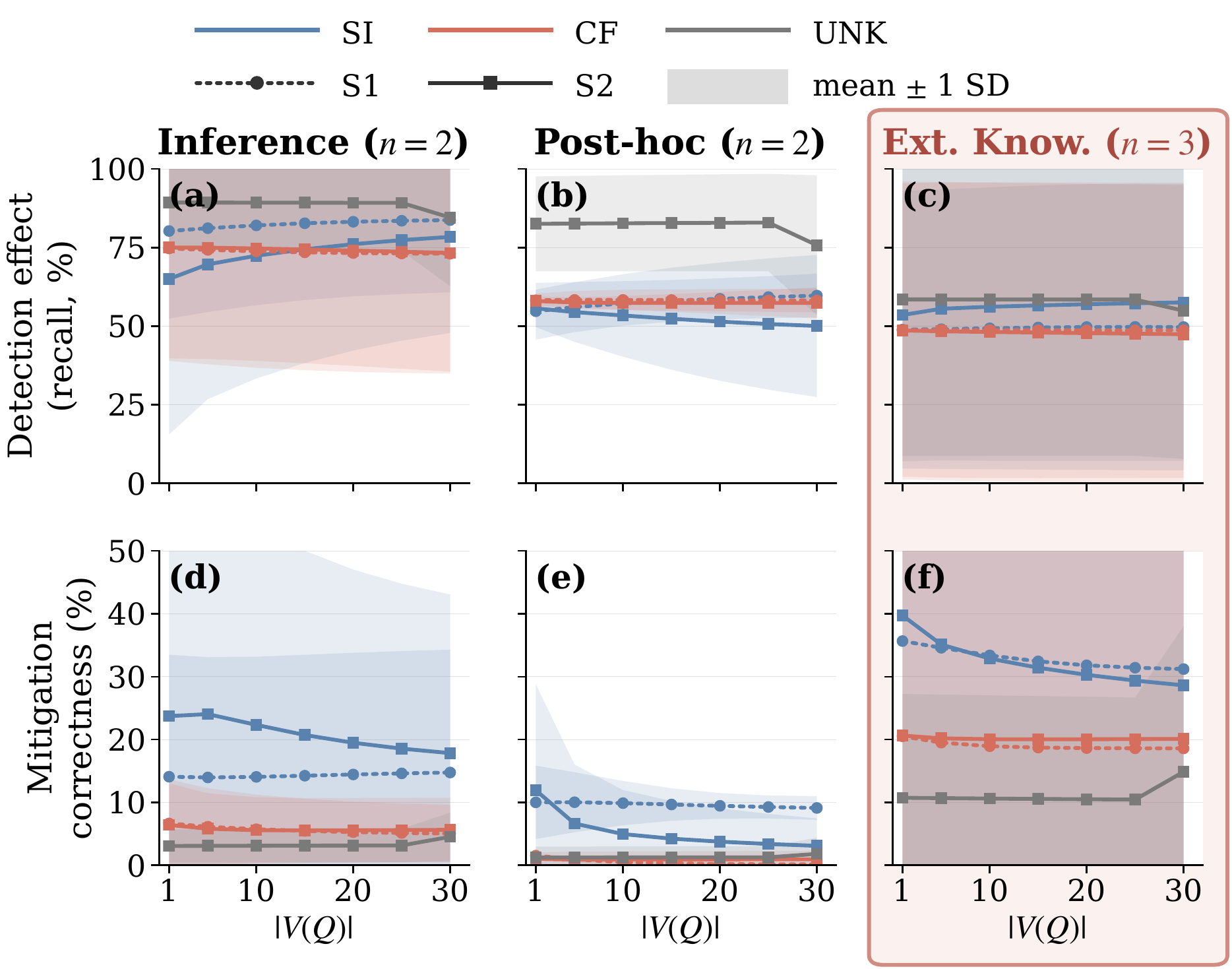}\\[-1pt]
    \caption{\footnotesize \textbf{Gaps by intervention stage}, as $|\VQ|$ grows. Only external-knowledge methods (highlighted) mitigate \CF{} well above zero (about 20\%), while inference-time and post-hoc methods stay near zero.}
  \label{fig:mitigation_gaps}
\end{figure}

\ecoparagraph{Where it comes from.} 
Each intervention stage shows a distinct pattern across almost every $|\VQ|$ from 1 to 30 (Fig.~\ref{fig:mitigation_gaps}).
Inference-time methods detect well but mitigate little, yet interestingly their \CF{} mitigation is not zero.
By acting on internal states (NoVo~\citep{ho2025novo}\footnote{For NoVo, the gain may also come from its candidate answers: it adds no facts and only ranks the given options, so we do not count it as an external-knowledge method, but a candidate set containing the correct answer can still supply it.} reads attention-head norms; ActLCD~\citep{zhang-etal-2025-active} contrasts layers), they may reach knowledge that prompting cannot, an open question for future work.
Post-hoc methods detect expressed uncertainty best but detect less overall than inference-time methods, and mitigate almost nothing beyond a few \SI{}.
External-knowledge methods are the only intervention whose \CF{} mitigation is clearly meaningful, with little decline as $|\VQ|$ grows; for \UNK{}, it even increases.
Yet they detect \SI{}, \CF{}, and \UNK{} at similar rates, so external knowledge is applied regardless of error type.
To make full use of it, methods should consider the error type to reduce unnecessary cost for the additional knowledge and assess whether the added information actually supplies the knowledge \CF{} is missing. (See \S\ref{subsec:intervention} for further analysis. )

\begin{takeaway}
\textbf{H2 is supported.}
Detecting \CF{} is more difficult than \SI{} both with and without uncertainty elicitation.
Mitigation is also asymmetric across all method types.
Even external-knowledge methods, which should supply the absent knowledge, mitigate \CF{} less than \SI{}.
Assessing whether the added knowledge is actually missing from the model could make them more effective.
Whether inference-time methods can reach knowledge that prompting cannot is an open question.
\end{takeaway}

\section{Finding 2: Uncertainty Estimation Does Not Always Work For Hallucinations}
\label{sec:f2}

Even though the majority of hallucination methods uniformly consider uncertainty estimation for hallucination detections or use it as a proxy (77\% in total, Fig.~\ref{fig:hallucination_funnels_main}), their performance has not been fully investigated. 
In this section, we explore which error types (\SI{}, \CF{}, and \UNK{}) these uncertainty estimation-based hallucination methods actually work for.

\ecoparagraph{Existing hallucination detection across method categories} 
We observe that \UNK{} responses are detected far more often than either hallucination type (\SI{} and \CF{}) across $|V(Q)|$ (Figs.~\ref{fig:mitigation_gaps} and~\ref{fig:f3}).

\begin{table}[h!]
\centering
\caption{\textbf{Uncertainty estimation-based hallucination detectors by label} (7 methods). For \UNK{}, a \emph{high}
score is a failure: these responses are not hallucinations. Mean $\pm$ std across methods.
}
\label{tab:uq}
\scriptsize
\setlength{\tabcolsep}{4pt}
\begin{tabular}{@{}l ccc ccc@{}}
\toprule
 & \multicolumn{3}{c}{S1} & \multicolumn{3}{c}{S2} \\
\cmidrule(lr){2-4}\cmidrule(lr){5-7}
 & \CF & \SI & \UNK & \CF & \SI & \UNK \\
\midrule
Hallucination score (mean) & 0.54{\scriptsize$\pm$.21} & 0.43{\scriptsize$\pm$.20} & \cellcolor{unkcol!15}\textbf{0.61}{\scriptsize$\pm$.25} & 0.42{\scriptsize$\pm$.29} & 0.42{\scriptsize$\pm$.21} & \cellcolor{unkcol!15}\textbf{0.57}{\scriptsize$\pm$.19} \\
Miss rate (\%) & 34.2{\scriptsize$\pm$25.6} & 35.6{\scriptsize$\pm$24.7} & 22.9{\scriptsize$\pm$21.4} & \cellcolor{cfcol!15}\textbf{50.4}{\scriptsize$\pm$33.5} & 35.9{\scriptsize$\pm$24.9} & 30.8{\scriptsize$\pm$23.0} \\
\midrule
Mitigation correct (\%) & 3.9{\scriptsize$\pm$7.3} & 29.5{\scriptsize$\pm$7.5} & 0.0{\scriptsize$\pm$0.0} & 2.2{\scriptsize$\pm$5.0} & 30.4{\scriptsize$\pm$8.6} & 0.4{\scriptsize$\pm$0.7} \\
\bottomrule
\end{tabular}
\end{table}

\ecoparagraph{Uncertainty estimation-based hallucination methods favor \UNK{}.} Under both schemes, uncertainty estimation-based hallucination methods give their highest mean score to \UNK{} (0.61 and 0.57), above \CF{} (0.54, 0.42) and \SI{} (0.43, 0.42), and miss \UNK{} least often (Table~\ref{tab:uq}; Fig.~\ref{fig:overview}c). 

\ecoparagraph{Uncertainty elicitation still does not help detect \CF{}.} Without uncertainty elicitation (S1), \CF{} and \SI{} are detected at similar rates; \CF{} is even missed slightly less often (34.2\% vs. 35.6\%).
Uncertainty elicitation (S2) moves most \CF{} into \UNK{} (the proportion of \CF{} falls from 0.36 to 0.04; Table~\ref{tab:labels}). 
The \CF{} that remain are those the model asserts even when allowed to say ``I don't know'', and detection methods miss half of them (50.4\%). 
This clearly shows that the \CF{} that remain are no easier to mitigate.

\begin{takeaway}
\textbf{H3 is not supported.} 
Existing hallucination methods detect \UNK{} more than \SI{} and \CF{}. In our definition, however, \UNK{} is not a hallucination. 
Uncertainty-based hallucination methods also score \UNK{} highest (0.61 / 0.57) and miss half of \CF{} under S2 (50.4\%). 
Report \UNK{} separately, or a method can be evaluated as good by simply detecting \UNK{}.
\end{takeaway}

We refer readers to \S\ref{sec:uncertainty} for further details in uncertainty-based methods for hallucinations.

\section{Finding 3: Only Latent Representations Work Meaningfully For Confabulations}
\label{sec:f3}

\begin{figure}[h!]
\vspace{-1em}
  \centering
  \includegraphics[width=0.65\linewidth]{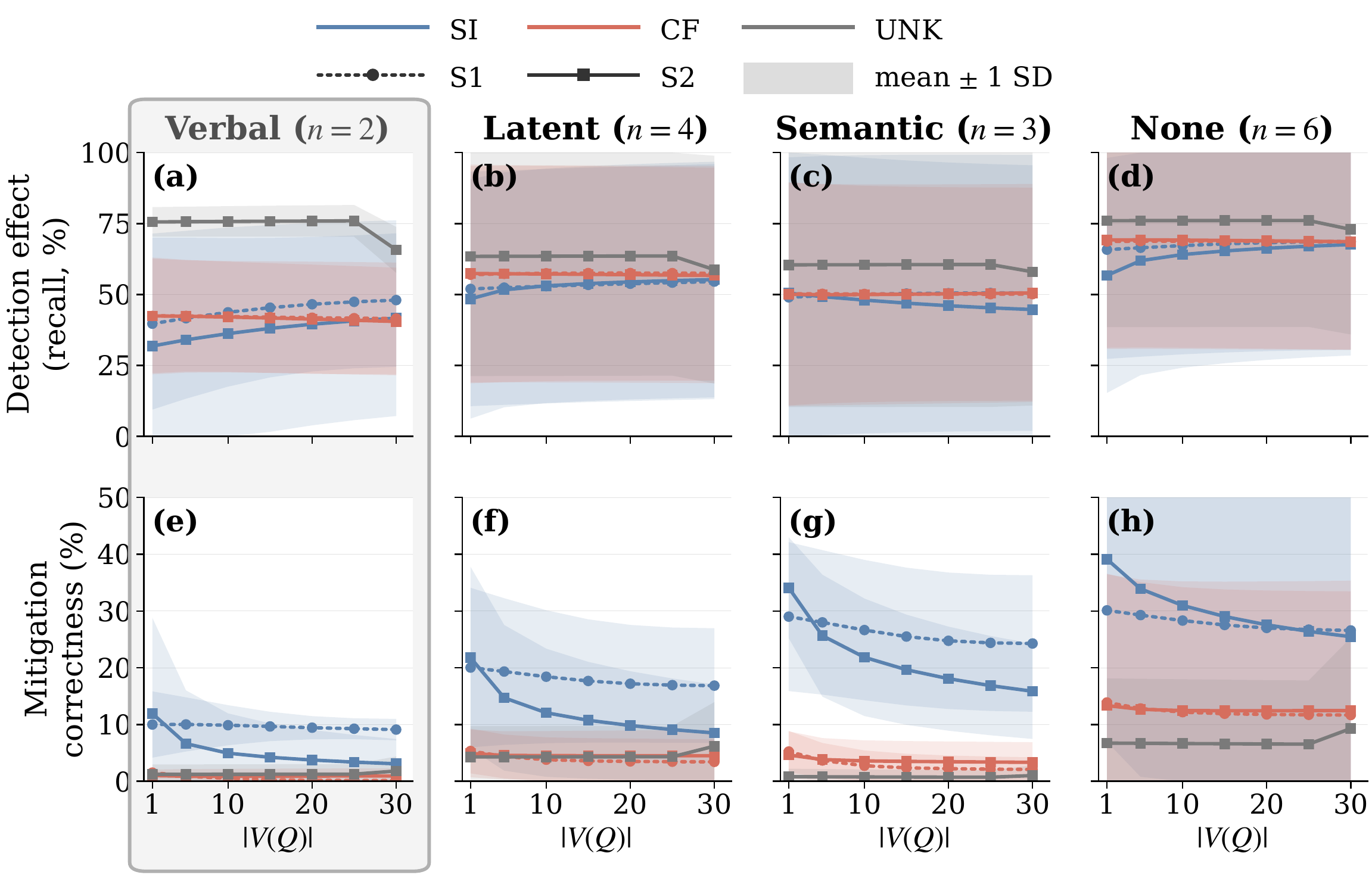}
  \caption{\textbf{Detection (top) and mitigation (bottom) by uncertainty representation} as $|\VQ|$
  grows. Verbal signals detect \UNK{} best and \CF{} worst; latent signals detect \CF{} better than
  \SI. $n$ is the number of methods per group; some methods belong to several groups.}
  \label{fig:f3}
  \vspace{-1em}
\end{figure}

Grouping uncertainty estimation-based hallucination methods by where they read uncertainty from (Fig.~\ref{fig:f3}):

\begin{itemize}[leftmargin=1.4em,itemsep=2pt,topsep=2pt]
  \item \textbf{Verbal} (output-level) signals give the best \UNK{} detection and the worst \CF{}
  detection. A \CF{} response contains no verbal cue to read.
  \item \textbf{Latent} (internal-state) signals give the best \CF{} detection among uncertainty
  representations, exceeding their own \SI{} detection at every $|\VQ|$, and the most \CF{} and \UNK{}
  mitigation among them.
  \item \textbf{Semantic} (consistency-based) signals give the best \SI{} mitigation, as expected:
  inconsistency across samples is exactly what defines \SI.
\end{itemize}

\noindent Methods that use no uncertainty at all mitigate \CF{} better still, because that group contains the external-knowledge methods. Detailed analysis is in \S\ref{sec:uncertainty}.

\begin{takeaway}
\textbf{H4 is supported.} 
Verbal uncertainty works best for \UNK{}, which states its uncertainty in the output, and semantic consistency works best for \SI{}. 
For \CF{}, which has no output-level cue, latent representations are the only uncertainty signal for which we observed a meaningful mitigation effect.
Since current methods succeed mostly on \SI{}, distinguishing error types is necessary to thoroughly understand existing methods and make meaningful progress on \CF{} and \UNK{} by matching the uncertainty representation to each type.
\end{takeaway}

%% file: sections/conclusion.tex
\section{Recommendations}
Based on our findings, we suggest four practical steps for hallucination research. First, \textbf{define hallucination as confident incorrectness} and explicitly distinguish it from expressed uncertainty, as conflating them risks targeting different phenomena under the same label. Second, \textbf{report \textsc{cf}, \textsc{si}, and \textsc{unk} separately} rather than aggregating all incorrect outputs, so that performance gains can be attributed to the right error type. Third, \textbf{test semantically equivalent phrasings before introducing external knowledge}: errors recoverable through rephrasing may not require retrieval augmentation, and this diagnostic can be often inexpensive relative to the interventions it can rule out. Fourth, \textbf{match uncertainty representation to the target error type}: verbal, latent, and semantic signals perform differently across each type, and latent representations are more promising than verbal ones for mitigating \textsc{cf} and \textsc{unk}. We hope these steps make hallucination evaluations more comparable across studies and help ensure that reported improvements reflect progress on the intended problem.

\section{Limitations and Conclusion}
Hallucination research currently lacks consistent operational distinctions across multiple error types. The term is applied inconsistently, and behaviorally different incorrect responses are often aggregated into a single metric. We decompose incorrect model outputs into three operational categories: confabulation (\textsc{cf}), semantic illusion (\textsc{si}), and expressed uncertainty (\textsc{unk}). 
Empirically, we show that \textsc{cf} remains substantially harder for existing methods to mitigate, and to detect under uncertainty elicitation, while \textsc{si} is often recoverable under inference-time intervention.
Although our study covers a limited set of models, datasets, and methods, the \textsc{si}--\textsc{cf} mitigation gap holds across both models, both labeling schemes, and all variant-set sizes.
Because \textsc{cf} is defined by persistent incorrectness under prompt variation, one might expect prompt-level methods to fail on \textsc{cf} by construction. This is not the case in our setup: $\mathcal{V}(Q)$ is used only for labeling, and evaluated methods see only the original prompt $c_0$, never the other conditions in $\mathcal{V}(Q)$ or the resulting labels.
The gap also holds for methods that do not use paraphrasing at all, including inference-time decoding and latent-state methods (Fig.~\ref{fig:comparison}c, \ref{fig:comparison_uncertainty}a), so the taxonomy predicts which interventions succeed rather than restating its definition.
We propose that future work define hallucination precisely, disaggregate error types, and audit alignment between definitions and evaluation methods.

\subsection*{AI use statement}
In this work, we used generative AI tools for generating synthetic data, specifically the semantically equivalent question paraphrases that make up $\mathcal{V}(Q)$, creating data visualization for our results, interpreting results, and refining and proof-reading the draft. 
We have reviewed all AI-assisted work.
Generated paraphrases were checked for semantic equivalence and for the absence of answer hints, and we further verified that they do not increase the likelihood of the gold answer.
AI-assisted interpretations were verified by the authors against the underlying experimental results and all text edits were reviewed by the authors.
We take responsibility for the final content of this work, including text, claims, or artifacts produced with the aid of generative AI.

\subsection*{Ethics statement}
Survey responses were collected from researchers who publish on LLM hallucinations. All responses were collected before respondents were informed of our hypotheses or taxonomy, to avoid biasing them. Participation was voluntary, all respondents agreed to the use of their responses, and only aggregate results are reported. All QA data come from publicly available benchmarks (TriviaQA, SimpleQA) and contain no private information.

\contributionbanner

%% file: sections/appendix.tex
\newpage
\section{Author Survey}
\label{app:survey}

We surveyed 24 first authors of hallucination papers published at top-tier
ML/NLP venues, asking which dimensions matter most for
categorizing hallucination (select all that apply). Responses were collected
before respondents saw our hypotheses or taxonomy.

\begin{figure}[h!]
    \centering
    \includegraphics[width=0.8\linewidth]{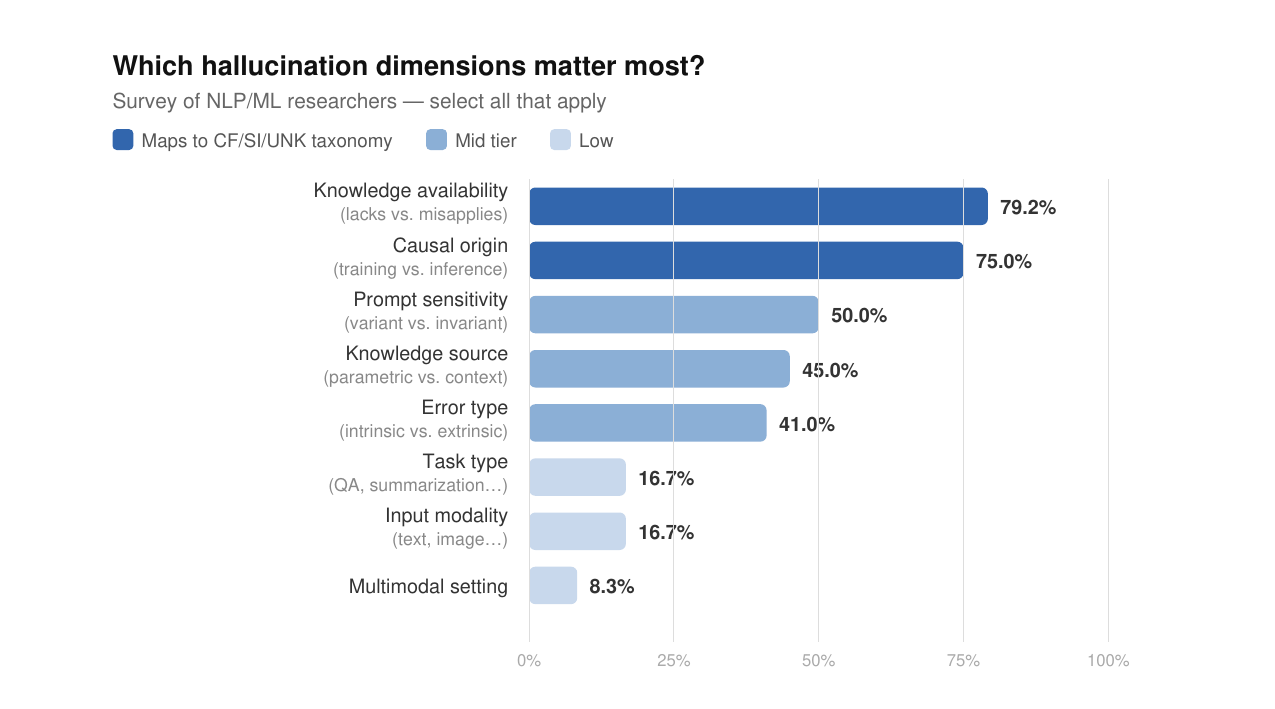}
    \caption{Dimensions that 24 first authors of hallucination papers consider
    most important for categorizing hallucination. Each bar is the fraction of
    respondents who selected that dimension; multiple selections were allowed,
    so percentages do not sum to 100\%. Dark bars mark the dimensions underlying
    our CF/SI/UNK taxonomy.}
    \label{fig:hallucination_survey_dimensions}
\end{figure}

As Figure~\ref{fig:hallucination_survey_dimensions} shows, the two most
frequently selected dimensions are \emph{knowledge availability}, whether the
model lacks the relevant knowledge or misapplies knowledge it has (79.2\%), and
\emph{causal origin}, whether the error arises from training data or from
inference-time dynamics (75.0\%). These are the axes along which our taxonomy
separates hallucinations (Section~\ref{sec:taxonomy}), so most researchers
independently favor the same distinctions. Dimensions concerning the knowledge
source, error type, and prompt sensitivity form a middle tier (41.0--50.0\%),
while those describing the setting rather than the cause of an error, namely task
type, input modality, and multimodality, were rarely selected (8.3--16.7\%).
Researchers thus view \emph{why} a model hallucinates, rather than \emph{where},
as the more useful basis for categorization.

\section{What the Community Measures}
\label{sec:community}

\begin{figure}[h!]
    \vspace{1em}
    \centering
    \includegraphics[width=\linewidth]{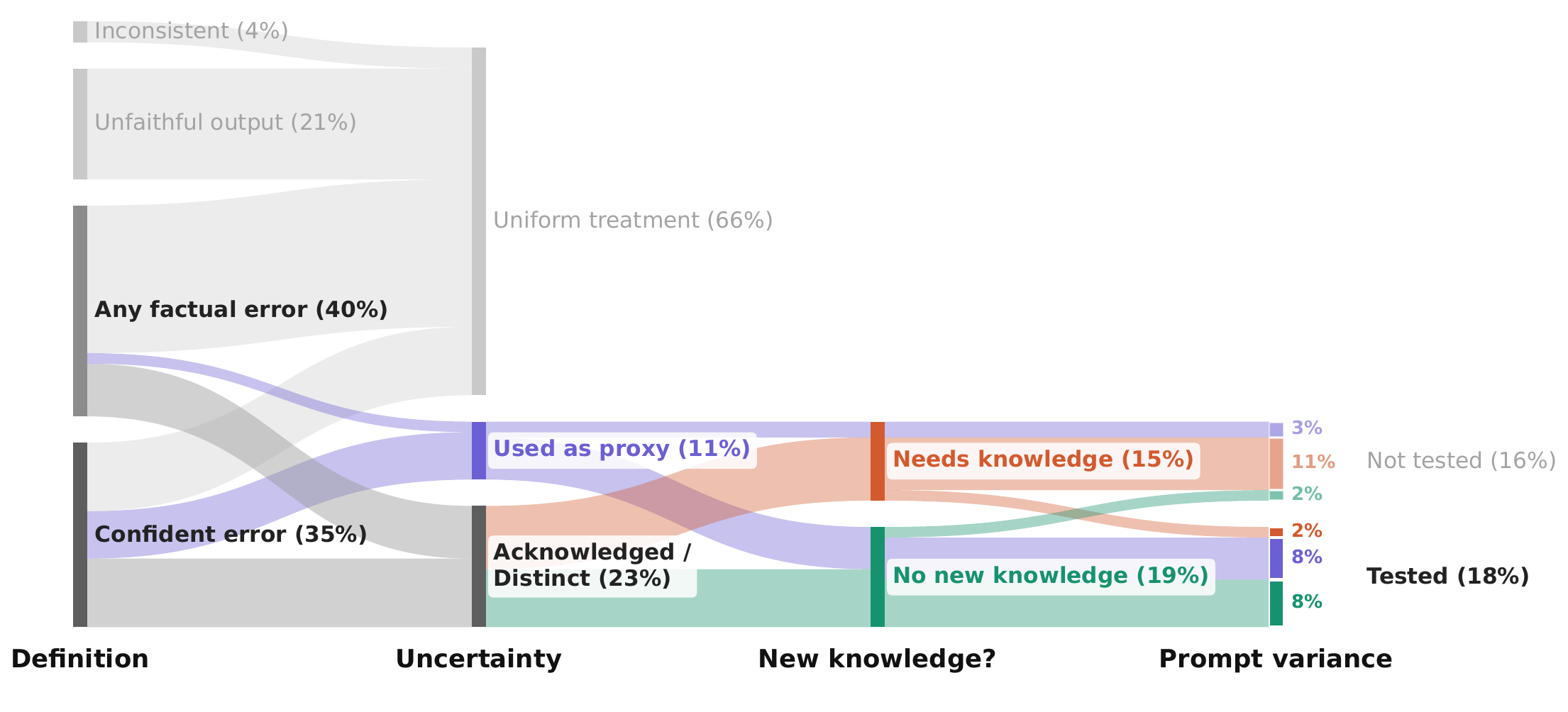}
    \vspace{-1em}
    \caption{\textbf{Recent text-based hallucination papers filtered by their definitions and approaches with uncertainty estimation, additional knowledge, and prompt variance.} From a pool of qualifying text-based papers$^{*}$ across ICML, NeurIPS, ICLR, ACL, and EMNLP (2021--2025), most papers neither distinguish hallucination from uncertainty nor test prompt variance before considering external knowledge as a source. The methods that skip these distinctions risk targeting the wrong error type entirely. Table~\ref{tab:text-papers} lists the papers with their category at each stage, and Table~\ref{tab:text-papers-ids} groups the paper IDs by category.}
    \label{fig:hallucination_funnels}
    \vspace{2pt}
    {\footnotesize\raggedright$^{*}$We did not filter papers further if they use the term ``hallucination'' in the title and take a text-based approach.\par}
    \vspace{1em}
\end{figure}

Recent hallucination-titled papers reveal two systematic methodological gaps (Fig.~\ref{fig:hallucination_funnels}).
The first is the hallucination-uncertainty\footnote{Note that uncertainty here encompasses both \textit{expressed uncertainty} and \textit{uncertainty estimation} methods, which are treated as distinct concepts throughout the paper.} distinction in definitions and methodologies. Only 35\% of papers nominally define hallucination as confident or plausible errors rather than \textit{expressed uncertainty}, and even this is not always reflected in methodology.
While 23\% treat hallucination detection as distinct from \textit{uncertainty estimation}\footnote{Methods that quantify epistemic uncertainty in model outputs, e.g., via token probabilities.}, not all anchor this in a confidence-based definition. 
Researchers broadly agree this distinction matters, as confirmed by a survey of 24 first authors of hallucination papers in top ML/NLP venues. 95.8\% believe hallucination should be distinguished from uncertainty (both expressed and estimated), yet only 16.7\% report treating them as distinct in their work.
We address this mismatch, a gap whose practical consequences we define in \S\ref{subsec:rq2} and quantify in \S\ref{sec:uncertainty}.

The second gap is knowledge diagnosis. 
Among papers engaging with hallucination and uncertainty, those concluding that new external knowledge is necessary (15\% of total) mostly never tested prompt variance first (11\% of total), despite many incorrect responses yielding correct ones under different question phrasings, instruction prompts, or inference-time decoding configurations (\S\ref{background:terminology} and \S\ref{subsec:distinction}).
Among those that do not introduce new knowledge (19\% of total), some never tested prompt variance either (3\% of total papers), potentially leaving their results suboptimal when simple prompt interventions could yield further gains. Without these diagnostics, methods risk over-engineering solutions to errors that simpler interventions would resolve.
This concern is crucial in hallucination research with 79.2\% of surveyed researchers identifying knowledge availability, whether a model lacks knowledge versus misapplies it, as the most critical dimension for characterizing hallucinations (Fig.~\ref{fig:hallucination_survey_dimensions}).

We surveyed the 91 text-based papers in our literature (Tables~\ref{tab:text-papers} and ~\ref{tab:text-papers-ids}) for the Figure~\ref{fig:hallucination_funnels}.

\textbf{Definition} (``How does the paper define hallucination?''): \textit{confident error} (``confident, factually incorrect output''; 35\%), \textit{any factual error} (``any factually incorrect output, regardless of expressed confidence''; 40\%), \textit{unfaithful output} (``output unfaithful to the provided context or document''; 21\%), or \textit{inconsistent output} (``output logically inconsistent within a single response''; 4\%).
Only the first two continue through the funnel.
\textbf{Uncertainty} (``Does the paper distinguish hallucination from uncertainty?''): \textit{acknowledged/distinct} if the paper answers ``yes, they are fundamentally different phenomena'' or ``partially, acknowledged but not operationalized''; \textit{used as proxy} if it answers ``no, it uses uncertainty signals as proxies''; and \textit{uniform treatment} if it answers ``no, it treats all incorrect outputs uniformly''. Of the 91 papers, 23\% are acknowledged/distinct, 11\% are used as proxy, and 66\% fall in the gray flow, which merges uniform treatment with papers whose definition fails the first filter.
\textbf{New knowledge} (``Are errors reducible without new knowledge?''): \textit{no new knowledge} if the paper reduces errors without new knowledge (19\%), and \textit{needs knowledge} if it requires external knowledge or retraining, or only partially fixes errors without it (15\%).
\textbf{Prompt variance} (``Do errors persist across reformulations of the prompt?''): \textit{tested} if the paper reports that errors are stable, disappear, or are mixed across paraphrases (18\%), and \textit{not tested} if it does not test this or the question is not applicable (16\%).
The last two stages are followed only for the 31 papers (34\%) that pass both of the first two filters.

{\footnotesize\setlength{\tabcolsep}{4pt}
\begin{longtable}{@{}r p{4.3cm} l l l l@{}}
\caption{All 91 text-modality papers in the survey. ``Needs knowledge'' includes ``partially''.}\label{tab:text-papers}\\
\toprule \# & Paper & Definition & Uncertainty & New knowledge & Prompt variance \\ \midrule \endfirsthead
\toprule \# & Paper & Definition & Uncertainty & New knowledge & Prompt variance \\ \midrule \endhead
\bottomrule \endfoot
1 & A Token-level Reference-free Hallucination Detection Benchmark~\citep{liu-etal-2022-token} & Any factual & Proxy & Needs knowledge & Not tested \\
2 & Hallucinated but Factual! Inspecting the Factuality of Hallucinations in Abstractive Summarization~\citep{cao-etal-2022-hallucinated} & Unfaithful & Ack./Distinct & Needs knowledge & Not tested \\
3 & Detecting and Mitigating Hallucinations in Machine Translation: Model Internal Workings~\citep{dale-etal-2023-detecting} & Unfaithful & Uniform & No new knowledge & Tested \\
4 & Optimal Transport for Unsupervised Hallucination Detection in Neural Machine Translation~\citep{guerreiro-etal-2023-optimal} & Unfaithful & Uniform & No new knowledge & Tested \\
5 & Self-Alignment for Factuality: Mitigating Hallucinations in LLMs via Self-Evaluation~\citep{zhang-etal-2024-self} & Confident & Ack./Distinct & Needs knowledge & Tested \\
6 & ANAH: Analytical Annotation of Hallucinations in Large Language Models~\citep{ji-etal-2024-anah} & Any factual & Uniform & Needs knowledge & Not tested \\
7 & TruthX: Alleviating Hallucinations by Editing Large Language Models in Truthful Space~\citep{zhang-etal-2024-truthx} & Confident & Ack./Distinct & No new knowledge & Tested \\
8 & InterrogateLLM: Zero-Resource Hallucination Detection in LLM-Generated Answers~\citep{yehuda-etal-2024-interrogatellm} & Confident & Ack./Distinct & No new knowledge & Not tested \\
9 & RAGTruth: A Hallucination Corpus for Developing Trustworthy Retrieval-Augmented Language Models~\citep{niu-etal-2024-ragtruth} & Unfaithful & Uniform & Needs knowledge & Not tested \\
10 & The Dawn After the Dark: An Empirical Study on Factuality Hallucination in Large Language Models~\citep{li-etal-2024-dawn} & Any factual & Uniform & Needs knowledge & Not tested \\
11 & Analyzing LLM Behavior in Dialogue Summarization: Unveiling Circumstantial Hallucination Trends~\citep{ramprasad-etal-2024-analyzing} & Unfaithful & Ack./Distinct & Needs knowledge & Not tested \\
12 & Confabulation: The Surprising Value of Large Language Model Hallucinations~\citep{sui-etal-2024-confabulation} & Any factual & Uniform & Needs knowledge & Not tested \\
13 & MPVStance: Mitigating Hallucinations in Stance Detection with Multi-Perspective Verification~\citep{zhang-etal-2025-mpvstance} & Any factual & Uniform & Needs knowledge & Not tested \\
14 & HALoGEN: Fantastic LLM Hallucinations and Where to Find Them~\citep{ravichander-etal-2025-halogen} & Any factual & Uniform & Needs knowledge & Not tested \\
15 & Hallucination Detox: Sensitivity Dropout (SenD) for Large Language Model Training~\citep{mohammadzadeh-etal-2025-hallucination} & Any factual & Proxy & Needs knowledge & Tested \\
16 & HD-NDEs: Neural Differential Equations for Hallucination Detection in LLMs~\citep{li-etal-2025-hd} & Any factual & Uniform & No new knowledge & Tested \\
17 & Beyond Facts: Evaluating Intent Hallucination in Large Language Models~\citep{hao-etal-2025-beyond} & Unfaithful & Ack./Distinct & Needs knowledge & Not tested \\
18 & DRAG: Distilling RAG for SLMs from LLMs~\citep{chen-etal-2025-drag} & Unfaithful & Uniform & Needs knowledge & Not tested \\
19 & Removal of Hallucination on Hallucination: Debate-Augmented RAG~\citep{hu-etal-2025-removal} & Unfaithful & Uniform & Needs knowledge & Not tested \\
20 & ICR Probe: Tracking Hidden State Dynamics for Reliable Hallucination Detection in LLMs~\citep{zhang-etal-2025-icr} & Confident & Uniform & No new knowledge & Tested \\
21 & Prompt-Guided Internal States for Hallucination Detection of Large Language Models~\citep{zhang-etal-2025-prompt} & Confident & Uniform & No new knowledge & Tested \\
22 & HalluLens: LLM Hallucination Benchmark~\citep{bang-etal-2025-hallulens} & Any factual & Ack./Distinct & Needs knowledge & Not tested \\
23 & Alleviating Hallucinations from Knowledge Misalignment in LLMs via Selective Abstention Learning~\citep{huang-etal-2025-alleviating} & Confident & Ack./Distinct & Needs knowledge & Not tested \\
24 & Stochastic Chameleons: Irrelevant Context Hallucinations Reveal Class-Based (Mis)Generalization in LLMs~\citep{cheng-etal-2025-stochastic} & Any factual & Uniform & No new knowledge & Tested \\
25 & TreeCut: A Synthetic Unanswerable Math Word Problem Dataset for LLM Hallucination Evaluation~\citep{ouyang-2025-treecut} & Confident & Uniform & Needs knowledge & Not tested \\
26 & Learning Auxiliary Tasks Improves Reference-Free Hallucination Detection in Open-Domain Long-Form Generation~\citep{qin-etal-2025-learning} & Any factual & Uniform & No new knowledge & Not tested \\
27 & Neural Path Hunter: Reducing Hallucination in Dialogue Systems via Path Grounding~\citep{dziri-etal-2021-neural} & Unfaithful & Uniform & Needs knowledge & Not tested \\
28 & Mutual Information Alleviates Hallucinations in Abstractive Summarization~\citep{van-der-poel-etal-2022-mutual} & Unfaithful & Proxy & No new knowledge & Not tested \\
29 & HalOmi: A Manually Annotated Benchmark for Multilingual Hallucination and Omission Detection in MT~\citep{dale-etal-2023-halomi} & Unfaithful & Uniform & Needs knowledge & Not tested \\
30 & Enhancing Uncertainty-Based Hallucination Detection with Stronger Focus~\citep{zhang-etal-2023-enhancing-uncertainty} & Confident & Ack./Distinct & No new knowledge & Tested \\
31 & The Troubling Emergence of Hallucination in Large Language Models~\citep{rawte-etal-2023-troubling} & Any factual & Ack./Distinct & Needs knowledge & Not tested \\
32 & Critic-Driven Decoding for Mitigating Hallucinations in Data-to-text Generation~\citep{lango-dusek-2023-critic} & Unfaithful & Uniform & No new knowledge & Not tested \\
33 & HaluEval: A Large-Scale Hallucination Evaluation Benchmark for Large Language Models~\citep{li-etal-2023-halueval} & Any factual & Uniform & Needs knowledge & Not tested \\
34 & Detecting and Mitigating Hallucinations in Multilingual Summarisation~\citep{qiu-etal-2023-detecting} & Unfaithful & Uniform & Needs knowledge & Not tested \\
35 & SelfCheckGPT: Zero-Resource Black-Box Hallucination Detection for Generative Large Language Models~\citep{manakul-etal-2023-selfcheckgpt} & Confident & Proxy & No new knowledge & Tested \\
36 & Hallucination Mitigation in Natural Language Generation from Large-Scale Open-Domain Knowledge Graphs~\citep{shi-etal-2023-hallucination} & Unfaithful & Uniform & Needs knowledge & Not tested \\
37 & KCTS: Knowledge-Constrained Tree Search Decoding with Token-Level Hallucination Detection~\citep{choi-etal-2023-kcts} & Unfaithful & Uniform & Needs knowledge & Not tested \\
38 & CRUSH4SQL: Collective Retrieval Using Schema Hallucination For Text2SQL~\citep{kothyari-etal-2023-crush4sql} & Any factual & Uniform & Needs knowledge & Not tested \\
39 & Hallucination Detection for Generative Large Language Models by Bayesian Sequential Estimation~\citep{wang-etal-2023-hallucination} & Confident & Ack./Distinct & Needs knowledge & Not tested \\
40 & Lookback Lens: Detecting and Mitigating Contextual Hallucinations in LLMs Using Only Attention Maps~\citep{chuang-etal-2024-lookback} & Unfaithful & Uniform & No new knowledge & Tested \\
41 & Embedding and Gradient Say Wrong: A White-Box Method for Hallucination Detection~\citep{hu-etal-2024-embedding} & Any factual & Uniform & No new knowledge & Tested \\
42 & Knowledge Verification to Nip Hallucination in the Bud~\citep{wan-etal-2024-knowledge} & Confident & Ack./Distinct & Needs knowledge & Not tested \\
43 & Whispers that Shake Foundations: Analyzing and Mitigating False Premise Hallucinations in LLMs~\citep{yuan-etal-2024-whispers} & Any factual & Uniform & No new knowledge & Tested \\
44 & Knowledge-Centric Hallucination Detection (RefChecker)~\citep{hu-etal-2024-knowledge} & Any factual & Uniform & Needs knowledge & Not tested \\
45 & ToolBeHonest: A Multi-level Hallucination Diagnostic Benchmark for Tool-Augmented LLMs~\citep{zhang-etal-2024-toolbehonest} & Any factual & Uniform & Needs knowledge & Not tested \\
46 & Null-Shot Prompting: Rethinking Prompting Large Language Models With Hallucination~\citep{taveekitworachai-etal-2024-null} & Any factual & Uniform & No new knowledge & Tested \\
47 & Small Agent Can Also Rock! Empowering Small Language Models as Hallucination Detector~\citep{cheng-etal-2024-small} & Any factual & Uniform & Needs knowledge & Not tested \\
48 & HalluMeasure: Fine-grained Hallucination Measurement Using Chain-of-Thought Reasoning~\citep{akbar-etal-2024-hallumeasure} & Any factual & Ack./Distinct & Needs knowledge & Not tested \\
49 & Enhanced Hallucination Detection in Neural Machine Translation through Simple Detector Aggregation~\citep{himmi-etal-2024-enhanced} & Unfaithful & Uniform & No new knowledge & Not tested \\
50 & MedHallu: A Comprehensive Benchmark for Detecting Medical Hallucinations in LLMs~\citep{pandit-etal-2025-medhallu} & Confident & Ack./Distinct & Needs knowledge & Not tested \\
51 & Active Layer-Contrastive Decoding Reduces Hallucination in Large Language Model Generation~\citep{zhang-etal-2025-active} & Any factual & Uniform & No new knowledge & Tested \\
52 & Calibrating Verbal Uncertainty as a Linear Feature to Reduce Hallucinations~\citep{ji-etal-2025-calibrating} & Confident & Ack./Distinct & No new knowledge & Tested \\
53 & The Impact of Negated Text on Hallucination with Large Language Models~\citep{seo-etal-2025-impact} & Any factual & Uniform & No new knowledge & Tested \\
54 & Attention-guided Self-reflection for Zero-shot Hallucination Detection in Large Language Models~\citep{liu-etal-2025-attention} & Confident & Ack./Distinct & No new knowledge & Tested \\
55 & Unsupervised Hallucination Detection by Inspecting Reasoning Processes~\citep{srey-etal-2025-unsupervised} & Confident & Proxy & No new knowledge & Tested \\
56 & Hallucination Detection in LLMs Using Spectral Features of Attention Maps~\citep{binkowski-etal-2025-hallucination} & Confident & Uniform & No new knowledge & Tested \\
57 & How Much Do LLMs Hallucinate across Languages? On Realistic Multilingual Estimation of LLM Hallucination~\citep{ul-islam-etal-2025-much} & Any factual & Uniform & Needs knowledge & Not tested \\
58 & Logit Space Constrained Fine-Tuning for Mitigating Hallucinations in LLM-Based Recommender Systems~\citep{deng-etal-2025-logit} & Any factual & Uniform & Needs knowledge & Not tested \\
59 & Detecting LLM Hallucination Through Layer-wise Information Deficiency~\citep{kim-etal-2025-detecting} & Confident & Proxy & No new knowledge & Tested \\
60 & The Illusion of Progress: Re-evaluating Hallucination Detection in LLMs~\citep{janiak-etal-2025-illusion} & Any factual & Uniform & Needs knowledge & Not tested \\
61 & A Head to Predict and a Head to Question: Pre-trained Uncertainty Quantification Heads for Hallucination Detection~\citep{shelmanov-etal-2025-head} & Confident & Ack./Distinct & No new knowledge & Tested \\
62 & Estimating the Hallucination Rate of Generative AI~\citep{jesson2024estimating} & Confident & Ack./Distinct & Needs knowledge & Not tested \\
63 & LLM-Check: Investigating Detection of Hallucinations in Large Language Models~\citep{sriramanan2024llmcheck} & Any factual & Uniform & No new knowledge & Not tested \\
64 & ERBench: An Entity-Relationship based Automatically Verifiable Hallucination Benchmark for LLMs~\citep{oh2024erbench} & Any factual & Uniform & Needs knowledge & Not tested \\
65 & ANAH-v2: Scaling Analytical Hallucination Annotation of Large Language Models~\citep{gu2024anahv} & Any factual & Uniform & Needs knowledge & Not tested \\
66 & HaloScope: Harnessing Unlabeled LLM Generations for Hallucination Detection~\citep{du2024haloscope} & Confident & Proxy & No new knowledge & Not tested \\
67 & Generalization or Hallucination? Understanding Out-of-Context Reasoning in Transformers~\citep{huang2025generalization} & Any factual & Ack./Distinct & Needs knowledge & Tested \\
68 & PHANTOM: A Benchmark for Hallucination Detection in Financial Long-Context QA~\citep{ji2025phantom} & Confident & Ack./Distinct & Needs knowledge & Not tested \\
69 & Beyond Token Probes: Hallucination Detection via Activation Tensors with ACT-ViT~\citep{bar-shalom2025beyond} & Confident & Uniform & No new knowledge & Tested \\
70 & SECA: Semantically Equivalent and Coherent Attacks for Eliciting LLM Hallucinations~\citep{liang2025seca} & Confident & Uniform & Needs knowledge & Tested \\
71 & Robust Hallucination Detection in LLMs via Adaptive Token Selection~\citep{niu2025robust} & Confident & Uniform & No new knowledge & Tested \\
72 & From Noise to Narrative: Tracing the Origins of Hallucinations in Transformers~\citep{suresh2025from} & Any factual & Uniform & No new knowledge & Tested \\
73 & FACT: Mitigating Inconsistent Hallucinations in LLMs via Fact-Driven Alternating Code-Text Training~\citep{you2025fact} & Inconsistent & Uniform & Needs knowledge & Tested \\
74 & Benford's Curse: Tracing Digit Bias to Numerical Hallucination in LLMs~\citep{shao_benfords_nodate} & Any factual & Uniform & No new knowledge & Tested \\
75 & Alleviating Hallucinations in LLMs through Multi-Model Contrastive Decoding and Dynamic Hallucination Detection~\citep{zhu2025alleviating} & Confident & Ack./Distinct & No new knowledge & Not tested \\
76 & Wisdom is Knowing What not to Say: Hallucination-Free LLMs Unlearning via Attention Shifting~\citep{tan2025wisdom} & Confident & Uniform & Needs knowledge & Not tested \\
77 & Auditing Meta-Cognitive Hallucinations in Reasoning Large Language Models~\citep{lu2025auditing} & Inconsistent & Ack./Distinct & Needs knowledge & Tested \\
78 & One SPACE to Rule Them All: Jointly Mitigating Factuality and Faithfulness Hallucinations in LLMs~\citep{wang2025one} & Any factual & Ack./Distinct & No new knowledge & Tested \\
79 & Self-contradictory Hallucinations of Large Language Models: Evaluation, Detection and Mitigation~\citep{mundler2024selfcontradictory} & Inconsistent & Uniform & No new knowledge & Tested \\
80 & INSIDE: LLMs' Internal States Retain the Power of Hallucination Detection~\citep{chen_inside_2024} & Confident & Proxy & No new knowledge & Tested \\
81 & Do I Know This Entity? Knowledge Awareness and Hallucinations in Language Models~\citep{ferrando2025do} & Confident & Ack./Distinct & No new knowledge & Tested \\
82 & ReDeEP: Detecting Hallucination in Retrieval-Augmented Generation via Mechanistic Interpretability~\citep{sun2025redeep} & Unfaithful & Uniform & No new knowledge & Tested \\
83 & Can Knowledge Editing Really Correct Hallucinations?~\citep{huang2025can} & Confident & Uniform & Needs knowledge & Not tested \\
84 & LLMs Know More Than They Show: On the Intrinsic Representation of LLM Hallucinations~\citep{orgad2025llms} & Any factual & Uniform & No new knowledge & Tested \\
85 & NoVo: Norm Voting off Hallucinations with Attention Heads in Large Language Models~\citep{ho2025novo} & Confident & Uniform & No new knowledge & Tested \\
86 & HaDeMiF: Hallucination Detection and Mitigation in Large Language Models~\citep{zhou2025hademif} & Any factual & Proxy & No new knowledge & Not tested \\
87 & How Language Model Hallucinations Can Snowball~\citep{ZhangPMLS24} & Inconsistent & Uniform & Needs knowledge & Tested \\
88 & In-Context Sharpness as Alerts: An Inner Representation Perspective for Hallucination Mitigation~\citep{chen2024incontext}{} & Confident & Proxy & No new knowledge & Tested \\
89 & Coarse-to-Fine Highlighting: Reducing Knowledge Hallucination in Large Language Models~\citep{lv2024coarsetofine} & Unfaithful & Uniform & Needs knowledge & Not tested \\
90 & Reducing Tool Hallucination via Reliability Alignment~\citep{xu2025reducing} & Any factual & Ack./Distinct & Needs knowledge & Not tested \\
91 & Steer LLM Latents for Hallucination Detection~\citep{park2025steer} & Confident & Proxy & No new knowledge & Tested \\
\end{longtable}}

\begin{table}[h]\centering\footnotesize
\caption{Paper IDs (as in Table~\ref{tab:text-papers}) by category, for all 91 text-modality papers.}\label{tab:text-papers-ids}
\begin{tabular}{@{}l l p{7cm}@{}}\toprule Stage & Category (n) & Paper IDs \\ \midrule
Definition & Confident error (32) & 5, 7, 8, 20, 21, 23, 25, 30, 35, 39, 42, 50, 52, 54, 55, 56, 59, 61, 62, 66, 68, 69, 70, 71, 75, 76, 80, 81, 83, 85, 88, 91 \\
 & Any factual error (36) & 1, 6, 10, 12, 13, 14, 15, 16, 22, 24, 26, 31, 33, 38, 41, 43, 44, 45, 46, 47, 48, 51, 53, 57, 58, 60, 63, 64, 65, 67, 72, 74, 78, 84, 86, 90 \\
 & Unfaithful output (19) & 2, 3, 4, 9, 11, 17, 18, 19, 27, 28, 29, 32, 34, 36, 37, 40, 49, 82, 89 \\
 & Inconsistent output (4) & 73, 77, 79, 87 \\ \midrule
Uncertainty & Acknowledged/Distinct (25) & 2, 5, 7, 8, 11, 17, 22, 23, 30, 31, 39, 42, 48, 50, 52, 54, 61, 62, 67, 68, 75, 77, 78, 81, 90 \\
 & Used as proxy (11) & 1, 15, 28, 35, 55, 59, 66, 80, 86, 88, 91 \\
 & Uniform treatment (55) & 3, 4, 6, 9, 10, 12, 13, 14, 16, 18, 19, 20, 21, 24, 25, 26, 27, 29, 32, 33, 34, 36, 37, 38, 40, 41, 43, 44, 45, 46, 47, 49, 51, 53, 56, 57, 58, 60, 63, 64, 65, 69, 70, 71, 72, 73, 74, 76, 79, 82, 83, 84, 85, 87, 89 \\ \midrule
New knowledge & No new knowledge (43) & 3, 4, 7, 8, 16, 20, 21, 24, 26, 28, 30, 32, 35, 40, 41, 43, 46, 49, 51, 52, 53, 54, 55, 56, 59, 61, 63, 66, 69, 71, 72, 74, 75, 78, 79, 80, 81, 82, 84, 85, 86, 88, 91 \\
 & Needs knowledge (48) & 1, 2, 5, 6, 9, 10, 11, 12, 13, 14, 15, 17, 18, 19, 22, 23, 25, 27, 29, 31, 33, 34, 36, 37, 38, 39, 42, 44, 45, 47, 48, 50, 57, 58, 60, 62, 64, 65, 67, 68, 70, 73, 76, 77, 83, 87, 89, 90 \\ \midrule
Prompt variance & Tested (41) & 3, 4, 5, 7, 15, 16, 20, 21, 24, 30, 35, 40, 41, 43, 46, 51, 52, 53, 54, 55, 56, 59, 61, 67, 69, 70, 71, 72, 73, 74, 77, 78, 79, 80, 81, 82, 84, 85, 87, 88, 91 \\
 & Not tested (50) & 1, 2, 6, 8, 9, 10, 11, 12, 13, 14, 17, 18, 19, 22, 23, 25, 26, 27, 28, 29, 31, 32, 33, 34, 36, 37, 38, 39, 42, 44, 45, 47, 48, 49, 50, 57, 58, 60, 62, 63, 64, 65, 66, 68, 75, 76, 83, 86, 89, 90 \\
\bottomrule\end{tabular}\end{table}

\clearpage
\newpage

\section{Definitions}
\label{app_sec:definitions}
\definecolor{cfcolor}{RGB}{250,243,240}   
\definecolor{sicolor}{RGB}{241,247,253}   
\definecolor{cfborder}{RGB}{200,150,140}  
\definecolor{siborder}{RGB}{145,175,205}  
\definecolor{headcf}{RGB}{140,40,15}      
\definecolor{headsi}{RGB}{15,70,140}      

\tcbset{
  defbox/.style 2 args={
    enhanced, breakable,
    colback=#1,
    colframe=#2,
    boxrule=0.8pt,
    arc=5pt,
    left=10pt, right=10pt, top=8pt, bottom=8pt,
    fonttitle=\bfseries\small\color{#2},
    attach boxed title to top left={yshift=-2.5mm, xshift=8pt},
    boxed title style={
      colback=white, colframe=#2,
      boxrule=0.8pt, arc=3pt,
      left=5pt, right=5pt, top=2pt, bottom=2pt
    }
  }
}

\vspace{1em}
\ecoparagraph{Notation.}
Let $c_0$ denote the original prompt, and let $\mathcal{V}(Q)$ be a set of semantically equivalent variants of question $Q$ (excluding $c_0$) sharing the same ground truth but introducing no additional hints. For each variant $v \in \{c_0\} \cup \mathcal{V}(Q)$, $r(v)$ denotes the model response; $\mathrm{correct}(r(v)) = 1$ if it matches the ground truth, $\mathrm{uncertain}(r(v)) = 1$ if the model expresses uncertainty (confidence score $= 0$ or explicit refusal).

\vspace{1em}
\begin{tcolorbox}[defbox={cfcolor}{cfborder},
                  title={Definition 1 \quad Confabulation (CF)},
                  fonttitle=\bfseries\small\color{headcf},
                  boxsep=4pt, top=6pt, bottom=1pt, left=5pt, right=5pt]
\linespread{1.02}\selectfont
\vspace{0.5em}
A response to question $Q$ is a \textbf{confabulation} if the model produces an incorrect, non-abstaining answer on \emph{every} variant in $\{c_0\} \cup \mathcal{V}(Q)$ under all plausible inference-time conditions (e.g., instruction variants, decoding configurations): \\[5pt]
\vspace{0.5em}
\hspace{6em}$\mathrm{CF}(Q)\iff
\forall v \in \{c_0\} \cup \mathcal{V}(Q):\;
\neg\,\mathrm{correct}(r(v)) \wedge
\neg\,\mathrm{uncertain}(r(v)).$
\end{tcolorbox}

\vspace{1em}
\begin{tcolorbox}[defbox={sicolor}{siborder},
                  title={Definition 2 \quad Semantic Illusion (SI)},
                  fonttitle=\bfseries\small\color{headsi},
                  boxsep=4pt, top=6pt, bottom=1pt, left=5pt, right=5pt]
\linespread{1.02}\selectfont
\vspace{0.5em}
A response to question $Q$ is a \textbf{semantic illusion} if the model answers the original prompt incorrectly but produces a correct answer on at least one variant, without any change to model parameters or the introduction of additional information:\\[5pt]
\vspace{0.5em}
\hspace{6em}$\mathrm{SI}(Q)\iff
\neg\,\mathrm{correct}(r(c_0)) \wedge
\exists v \in \mathcal{V}(Q):\;
\mathrm{correct}(r(v)).$

\end{tcolorbox}
\medskip

\vspace{1em}
\begin{tcolorbox}[defbox={gray!5}{gray!70},
                  title={Definition 3 \quad Expressed Uncertainty (\UNK{})},
                  fonttitle=\bfseries\small\color{gray!60!black},
                  boxsep=4pt, top=6pt, bottom=1pt, left=5pt, right=5pt]
\linespread{1.02}\selectfont
\vspace{0.5em}
   A response to question $Q$ is an \textbf{expressed uncertainty} if the model produces no correct answer on any variant in $\{c_0\} \cup \mathcal{V}(Q)$ but verbally expresses uncertainty on at least one.
\vspace{0.5em}
\end{tcolorbox}

\vspace{2em}
\begin{remark2}[Stability relative to $\mathcal{V}(Q)$]
\label{remark}
\CF{} and \SI{} are defined with respect to a finite evaluation set $\mathcal{V}(Q)$.
Expanding $\mathcal{V}(Q)$ may reclassify some \CF{} instances as \SI{} if a correct response is elicited.
We therefore treat them as \emph{operational diagnostics}, not intrinsic model properties. 
This reflects practical constraints. 
Model behavior can only be probed under a limited set of prompts and decoding settings.
Under this constraint, \CF{} corresponds to cases where no tested condition yields a correct response, while \SI{} corresponds to errors recoverable within the same evaluation budget.
Although labels may change as $\mathcal{V}(Q)$ grows, the mitigation gap between \CF{} and \SI{} remains consistent across schemes.
We show the robust empirical observations despite different $\mathcal{V}(Q)$ in \S\ref{sec:spectrum}.
\end{remark2}

\newpage
\section{Methods}
\label{sec:methods}

\paragraph{Study Design}

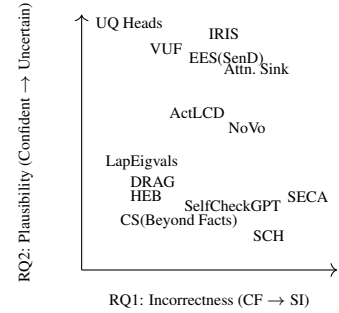
\begin{wrapfigure}{r}{0.33\textwidth}
\vspace{-1em}
\centering
\begin{tikzpicture}[scale=0.65]
\begin{scope}[xshift=0.8cm]
\draw[->] (0,0) -- (5.2,0);
\draw[->] (0,0) -- (0,5.2);
\node[anchor=north] at (2.6, -0.3) {\tiny RQ1: Incorrectness (CF $\rightarrow$ SI)};
\node[rotate=90, anchor=south] at (-0.8, 2.6) {\tiny RQ2: Plausibility (Confident $\rightarrow$ Uncertain)};

\node[anchor=west] at (0.1, 5.0) {\tiny UQ Heads};
\node[anchor=west] at (1.2, 4.5) {\tiny VUF};
\node[anchor=west] at (2.4, 4.8) {\tiny IRIS};
\node[anchor=west] at (2.0, 4.3) {\tiny EES(SenD)};
\node[anchor=west] at (2.7, 4.1) {\tiny Attn.\ Sink};
\node[anchor=west] at (1.6, 3.2) {\tiny ActLCD};
\node[anchor=west] at (2.8, 2.9) {\tiny NoVo};

\node[anchor=west] at (0.3, 2.2) {\tiny LapEigvals};
\node[anchor=west] at (0.8, 1.8) {\tiny DRAG};
\node[anchor=west] at (0.8, 1.5) {\tiny HEB};
\node[anchor=west] at (0.6, 1.0) {\tiny CS(Beyond Facts)}; 

\node[anchor=west] at (1.9, 1.3) {\tiny SelfCheckGPT};
\node[anchor=west] at (3.3, 0.7) {\tiny SCH};
\node[anchor=west] at (4.0, 1.5) {\tiny SECA};

\end{scope}
\end{tikzpicture}
\caption{\textbf{Methods in RQ1--RQ2 space.}}
\label{fig:rq-map}

\end{wrapfigure}

We analyze recent hallucination research and conduct a controlled empirical evaluation. 
We collect 14 papers (2023--2025) from major ML and NLP venues with ``hallucination'' in the title, and categorize them based on methodological choices reported by the authors.
Dimensions are organized around our two research questions, as summarized in Table~\ref{tab:rq-summary} and visualized in Figure~\ref{fig:rq-map}. For RQ1, we examine prompt sensitivity (whether prompt variation is modeled), intervention stage (inference-time vs. post-hoc vs. external knowledge), and model knowledge assumption (whether the method implicitly assumes knowledge is absent or present). For RQ2, we examine the usage of uncertainty estimation (whether uncertainty is used for calibration or correction) and uncertainty representation (verbal, latent, semantic, or none). 
Together, these dimensions position existing methods within our taxonomy and show how their design choices relate to different error types.

\begin{table*}[h!]
\centering
\scriptsize
\vspace{1em}
\renewcommand{\arraystretch}{0.72}
\caption{\textbf{Experimental dimensions for RQ1--RQ2.} H1: prompt sensitivity, intervention stage, and target error type; H2: knowledge assumption; H3--H4: uncertainty usage and representation. Dimensions are explained in \S\ref{app_dimensions_notation}.}
\vspace{-1em}
\begin{tabular}{@{}p{0.25cm} p{0.25cm} p{2.2cm} p{1.4cm} p{8.5cm}@{}}
\toprule
\textbf{RQ} & \textbf{H} & \textbf{Angle} & \textbf{Category} & \textbf{Methods} \\
\midrule
RQ1 & H1 & Prompt Sensitivity
& Modeled & SECA~\cite{liang2025seca}, SelfCheckGPT~\cite{manakul-etal-2023-selfcheckgpt} \\
& & & Not modeled & DRAG~\cite{chen-etal-2025-drag}, UQ Heads~\cite{shelmanov-etal-2025-head}, ActLCD~\cite{zhang-etal-2025-active} \\
\midrule
RQ1 & H1 & Intervention Stage
& Inference & ActLCD~\cite{zhang-etal-2025-active}, NoVo~\cite{ho2025novo} \\
& & & Post-hoc & SelfCheckGPT~\cite{manakul-etal-2023-selfcheckgpt}, UQ Heads~\cite{shelmanov-etal-2025-head} \\
& & & Ext.\ Know. & DRAG~\cite{chen-etal-2025-drag}, HEB~\cite{huang2025can}, CS(Beyond Fact)~\cite{hao-etal-2025-beyond} \\
\midrule
RQ1 & H2 & Knowledge Assump.
& Absent & SelfCheckGPT~\cite{manakul-etal-2023-selfcheckgpt}, IRIS~\cite{srey-etal-2025-unsupervised}, VUF~\cite{ji-etal-2025-calibrating}, SCH~\cite{mundler2024selfcontradictory}, EES(SenD)~\cite{mohammadzadeh-etal-2025-hallucination}, Attn.\ Sink~\cite{binkowski2026attentionsinksinternalsignals} \\
& & & Present & NoVo~\cite{ho2025novo}, ActLCD~\cite{zhang-etal-2025-active}, LapEigvals~\cite{binkowski-etal-2025-hallucination}, SECA~\cite{liang2025seca} \\
\midrule
RQ2 & H3 & Uncertainty Usage
& Calibration & VUF~\cite{ji-etal-2025-calibrating}, UQ Heads~\cite{shelmanov-etal-2025-head}, IRIS~\cite{srey-etal-2025-unsupervised}, Attn.\ Sink~\cite{binkowski2026attentionsinksinternalsignals}, SelfCheckGPT~\cite{manakul-etal-2023-selfcheckgpt}  \\
& & & Correction & EES(SenD)~\cite{mohammadzadeh-etal-2025-hallucination}, SCH~\cite{mundler2024selfcontradictory} \\
\midrule
RQ2 & H4 & Uncertainty Repr.
& Verbal & VUF~\cite{ji-etal-2025-calibrating}, UQ Heads~\cite{shelmanov-etal-2025-head} \\
& & & Latent & UQ Heads~\cite{shelmanov-etal-2025-head}, EES(SenD)~\cite{mohammadzadeh-etal-2025-hallucination}, IRIS~\cite{srey-etal-2025-unsupervised}, Attn.\ Sink~\cite{binkowski2026attentionsinksinternalsignals} \\
& & & Semantic & SelfCheckGPT~\cite{manakul-etal-2023-selfcheckgpt}, SCH~\cite{mundler2024selfcontradictory}, IRIS~\cite{srey-etal-2025-unsupervised} \\
& & & None & SECA~\cite{liang2025seca}, DRAG~\cite{chen-etal-2025-drag}, HEB~\cite{huang2025can}, NoVo~\cite{ho2025novo}, ActLCD~\cite{zhang-etal-2025-active}, LapEigvals~\cite{binkowski-etal-2025-hallucination} \\
\bottomrule
\end{tabular}
\label{tab:rq-summary}
\end{table*}

\newpage
\subsection{Taxonomy Dimensions}
\label{app_dimensions_notation}
\paragraph{Prompt sensitivity}
A method models prompt sensitivity if it explicitly evaluates how responses vary across semantically equivalent prompt variants $\mathcal{V}(Q)$. Methods testing on a single prompt do not model this variation.

\paragraph{Intervention stage}
We distinguish three points where a method can intervene: during generation (\textit{inference}), after generation (\textit{post-hoc}), or by augmenting the model capabilities with \textit{external knowledge}.

\paragraph{Knowledge assumption}
A method either assumes the correct knowledge exists in the model but is misapplied, pointing to a retrieval failure (\textit{present knowledge}), or assumes the knowledge is absent entirely and must be supplied through external sources (i.e., external knowledge), pointing to a knowledge gap (\textit{absent knowledge}).

\paragraph{Uncertainty usage}
Uncertainty signals can serve two purposes, either (i) estimating how likely a response is to be correct (\textit{calibration}), or (ii) actively modifying outputs based on that signal (\textit{correction})

\paragraph{Uncertainty representation}
A method may capture, express, or quantify uncertainty through expressed uncertainty in outputs (\textit{verbal}), internal model states (\textit{latent}), consistency across multiple outputs (\textit{semantic}), or not engage with uncertainty at all (\textit{none}).

\paragraph{Target error}
Each method primarily targets an error type from our taxonomy: \textit{semantic illusion}, addressed through prompt sensitivity modeling; \textit{confabulation}, addressed through knowledge augmentation; or \textit{expressed uncertainty}, addressed through verbal uncertainty signals.

\newpage
\section{Dataset Construction}
\label{app:dataset}

This appendix provides the complete dataset construction pipeline, including prompt templates, behavioral grid specification, and labeling criteria.
The main paper (\S\ref{sec:setup}) summarizes and selectively presents the design; this appendix provides the detail needed for replication.

\begin{wraptable}{r}{0.3\linewidth}
    \vspace{-1em}
    \centering
    \scriptsize
    \caption{Label distribution}
    \setlength{\tabcolsep}{3pt}
    \renewcommand{\arraystretch}{1.05}
    \begin{tabular}{lccc}
    \toprule
    Scheme & CF & SI & UNK \\
    \midrule
    \textbf{S1} & 0.36 & 0.62 & 0.02 \\
    \textbf{S2} & 0.04 & 0.64 & 0.32 \\
    \bottomrule
    \end{tabular}
    \label{fig:label_combined}
\end{wraptable}
We construct a controlled factual QA dataset drawn from TriviaQA~\cite{joshi2017triviaqa} and SimpleQA~\cite{wei2024measuring} across (\textsc{nh},\SI{},\CF{},\UNK{}) categories for our study.
We study closed-book factual question answering, because it provides verifiable ground truth, makes model knowledge the sole explanatory variable, and removes retrieval or grounding confounds. 
We assign them by 30 semantically equivalent question variants for a practical evaluation.
We report all main results across two different instruction methods (see details in \S\ref{app:dataset}).
\textbf{S1} applies a richer grid of 30 semantically equivalent variants, 3 instruction prompts without uncertainty elicitation, and multiple inference-time decoding configurations (e.g., temperatures, see details in \S\ref{app:dataset:grid}); \textbf{S2} additionally includes an uncertainty-elicitation instruction.

\subsection{Source Datasets}
\label{app:dataset:sources}

Candidate questions are drawn from two publicly available factual QA
benchmarks:

\begin{itemize}
  \item \textbf{TriviaQA}~\cite{joshi2017triviaqa}: open-domain trivia questions
    with verified answers. We use the unfiltered development split and
    restrict to questions with short-phrase answers
    to ensure unambiguous correctness evaluation.
  \item \textbf{SimpleQA}~\cite{wei2024measuring}: a benchmark specifically
    designed to probe factual knowledge with short, verifiable questions
    and a structured confidence-elicitation protocol.
    We adopt SimpleQA's calibration prompt (see
    \S\ref{app:dataset:prompt}).
\end{itemize}

\noindent
We apply the following pre-filters before behavioral evaluation:
(i) answers must be short and unambiguous (no date ranges, no
``either X or Y'' answers);
(ii) questions must not require world-knowledge updated after the training cutoff of any construction model for semantic illusion type.
The remaining pool of candidates is passed to the behavioral evaluation grid described in \S\ref{app:dataset:grid}.

\subsection{Confidence Prompt Template}
\label{app:dataset:prompt}

All responses during construction and evaluation are elicited using
the following prompt, adapted directly from SimpleQA's calibration method:

\begin{tcolorbox}[
  arc=0.1mm,
  sharp corners,
  colframe=black!80,
  boxrule=0.8pt,
  title=\textbf{Confidence Prompt (used for all models)},
  fonttitle=\small\bfseries,
  coltitle=white,
  fontupper=\small\ttfamily,
  left=6pt, right=6pt, top=4pt, bottom=4pt
]
Here is the question: \{question\}\\[4pt]
Please provide your best guess and a confidence score between 0\%
to 100\% in the following JSON format:\\[4pt]
\{\\
\quad "answer": "Your answer here",\\
\quad "confidence\_score": number\\
\}
\end{tcolorbox}

\begin{figure}
    \centering
    \includegraphics[width=0.9\linewidth]{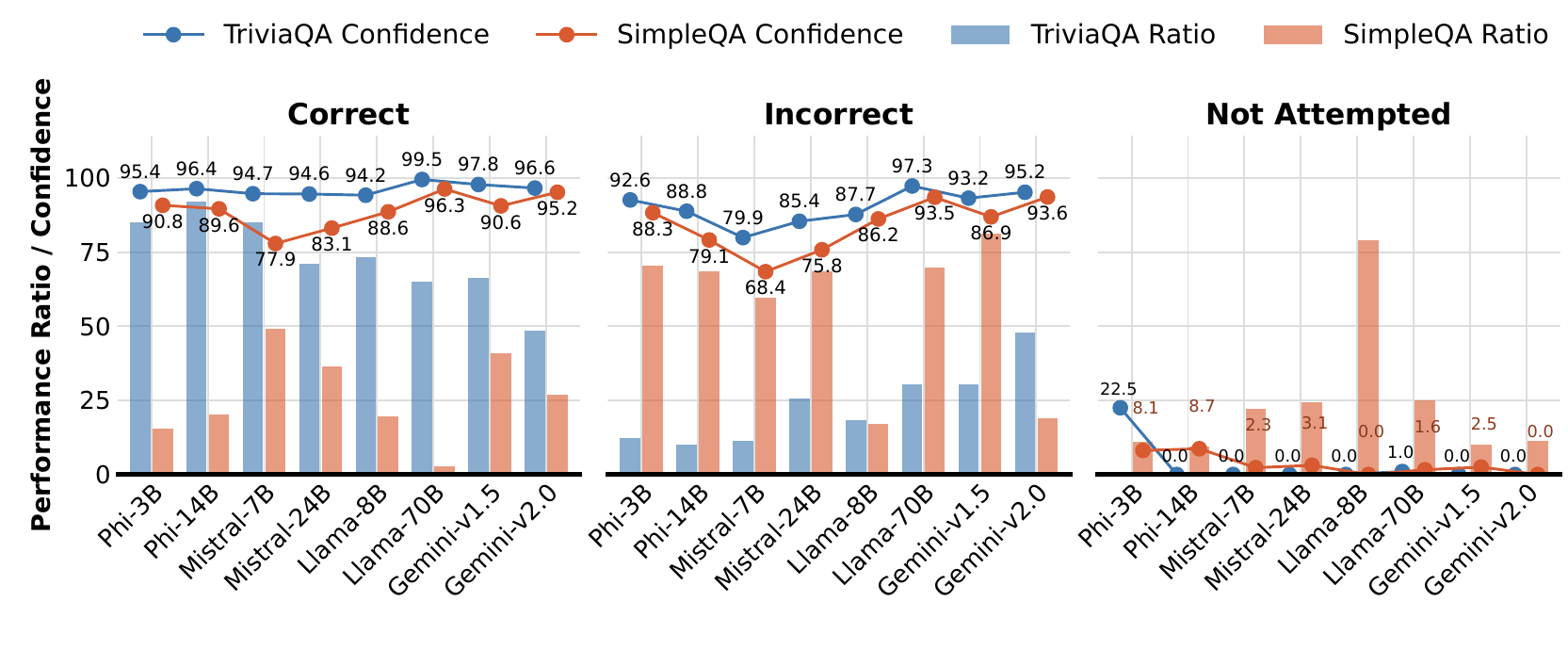}
    \caption{Model confidence scores across different response categories -- correct/incorrect responses and not attempted -- on the TriviaQA validation split (5k) and SimpleQA (4.3k). High confidence on incorrect responses and near-zero confidence on not-attempted responses show that the confidence score is consistent with explicit verbal uncertainty.}
    \label{fig:output_uncertainty}
\end{figure}

\noindent
We observe that under this prompt, models across families consistently assign high confidence to incorrect responses (Figure~\ref{fig:output_uncertainty}).
When models explicitly state that they do not know or refuse to answer, their confidence scores are almost always near zero~\cite{wei2024measuring}.
The confidence score therefore adds little information beyond the verbal expression of uncertainty.

\subsection{Behavioral Evaluation}
\label{app:dataset:grid}

Each candidate questions per category \textsc{nh}, \SI{}, \CF{}, and \UNK{} are evaluated under a dense behavioral grid designed to probe response stability across input perturbations that are semantically equivalent but surface-form variable.

\paragraph{Instruction variants.}
We evaluate each question under 4 instruction prompt variants that vary the framing of the task.
Each instruction variant is crossed with each perturbation condition.

\begin{tcolorbox}[
  sharp corners,
  colback=black!3,
  colframe=black!80,
  boxrule=0.8pt,
  title=\textbf{Instruction Prompt Variants},
  fonttitle=\small\bfseries,
  coltitle=white,
  fontupper=\small\ttfamily,
  left=6pt, right=6pt, top=4pt, bottom=4pt
]
\small
\begin{tabular}{@{}lp{12.5cm}@{}}
  \textbf{1} & \texttt{Answer the question with a short factual phrase. Output only the answer.} \\[4pt]
  \textbf{2} & \texttt{Return a concise answer only. No explanation.} \\[4pt]
  \textbf{3} & \texttt{Provide only the entity/date/name that answers the question.} \\[4pt]
  \textbf{4} & \texttt{Give the most likely factual answer. If uncertain, say: I don't know.} \normalfont\textit{(uncertainty-elicitation)} \\
\end{tabular}
\end{tcolorbox}

\paragraph{Inference-time decoding configurations.}
We evaluate each (question, condition, instruction) triple under multiple decoding settings: 
greedy decoding, temperature sampling ($T \in \{0.2, 0.7\}$), and nucleus sampling ($p = 0.80, 0.95$).
This ensures that behavioral labels are not artifacts of a single decoding strategy.

\paragraph{Semantically equivalent paraphrase variants.}
We additionally evaluate each question under 30 semantically equivalent paraphrases that vary linguistic structure and lexical choices without adding hints or information.

\newpage
\section{What is Accessible Knowledge?}
\label{app_sec:accessible_knowledge}

We define knowledge as \emph{accessible} if at least one variant $v \in \mathcal{V}(Q)$ elicits a correct response without additional hints or parameter changes.  
Under this definition, \SI{} reflects correctness that is elicitable under some semantically equivalent variants but not others, while \CF{} reflects persistent incorrectness across all the variants in $\mathcal{V}(Q)$.
The rest of this section validates that this behavioral distinction is stable, surprisingly inexpensive, and measurable in practice.

\paragraph{$\rho_k$ as a behavioral (knowledge) accessibility measure.}
For a question $Q$, the correctness proportion
\[\rho_k(Q) = \frac{1}{|\mathcal{V}_k(Q)|} \sum_{v \in \mathcal{V}_k(Q)} \mathbf{1}[\mathrm{correct}(r(v))]\]
estimates the probability that a uniformly drawn variant elicits a correct response, serving as a behavioral measure of response accessibility under perturbation.
The four categories correspond to distinct regimes:
\[
\rho_k \approx 0 \;\Rightarrow\; \CF{},\UNK{}, \qquad
0 < \rho_k < 1 \;\Rightarrow\; \SI{}, \qquad
\rho_k \approx 1 \;\Rightarrow\; \textsc{nh}.
\]
Both \CF{} and \UNK{} have $\rho_k \approx 0$ although they are behaviorally distinct as  \CF{} is confidently wrong while \UNK{} explicitly refuses to answer or expresses uncertainty.

Let $H_{\mathrm{bin}}(\rho) = -\rho\log_2\rho - (1-\rho)\log_2(1-\rho)$, which is maximized (1 bit) at $\rho{=}0.5$ and approaches zero at both extremes. 
\CF{} and \UNK{} questions have $\rho_k = 0$ and hence $H_{\mathrm{bin}} = 0$.
\SI{} questions have $\rho_k > 0$ by construction, but their correctness is typically low: the median $\rho_{30}$ is 0.17 (Gemma) and 0.13 (Llama), so a correct answer is elicitable but appears on only a minority of variants. Accordingly, $H_{\mathrm{bin}}$ for \SI{} plateaus around 0.5--0.6 bits once $|\mathcal{V}(Q)| \ge 10$ (Fig.~\ref{fig:vq}), the highest among all categories. Correct responses can thus be elicited but remain behaviorally unstable across semantically equivalent variants, a pattern we refer to as accessible knowledge.

\section{Results: Detailed  Analysis}
\label{sec:results}

Across all methodological dimensions, models, and labeling schemes, one asymmetry is consistent: \SI{} is both detectable and mitigable, whereas \CF{} is rarely mitigated and, under S2, less often detected.
In this section, we empirically study that this is a property of \CF{} itself, rather than of any individual method.

\subsection{Intervention Asymmetry}
\label{subsec:intervention}

\begin{figure}[h!]
    \centering
    \begin{subfigure}[t]{0.4\linewidth}
        \centering
        \includegraphics[width=\linewidth]{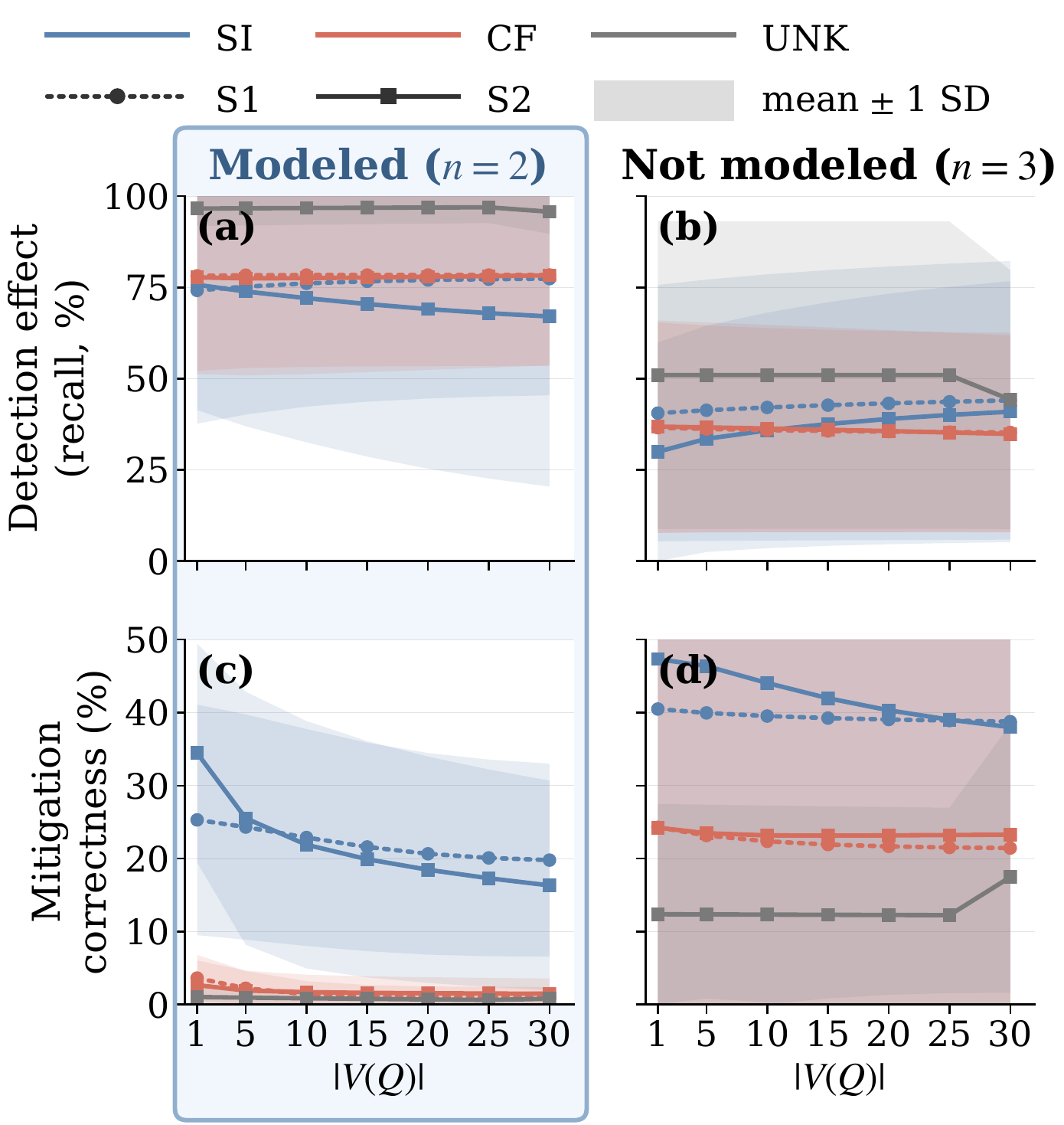}
        \caption{Prompt Sensitivity}
        \label{fig:comparison_prompt_sensitivity}
    \end{subfigure}
    \hspace{1em}
    \begin{subfigure}[t]{0.4\linewidth}
        \centering
        \includegraphics[width=\linewidth]{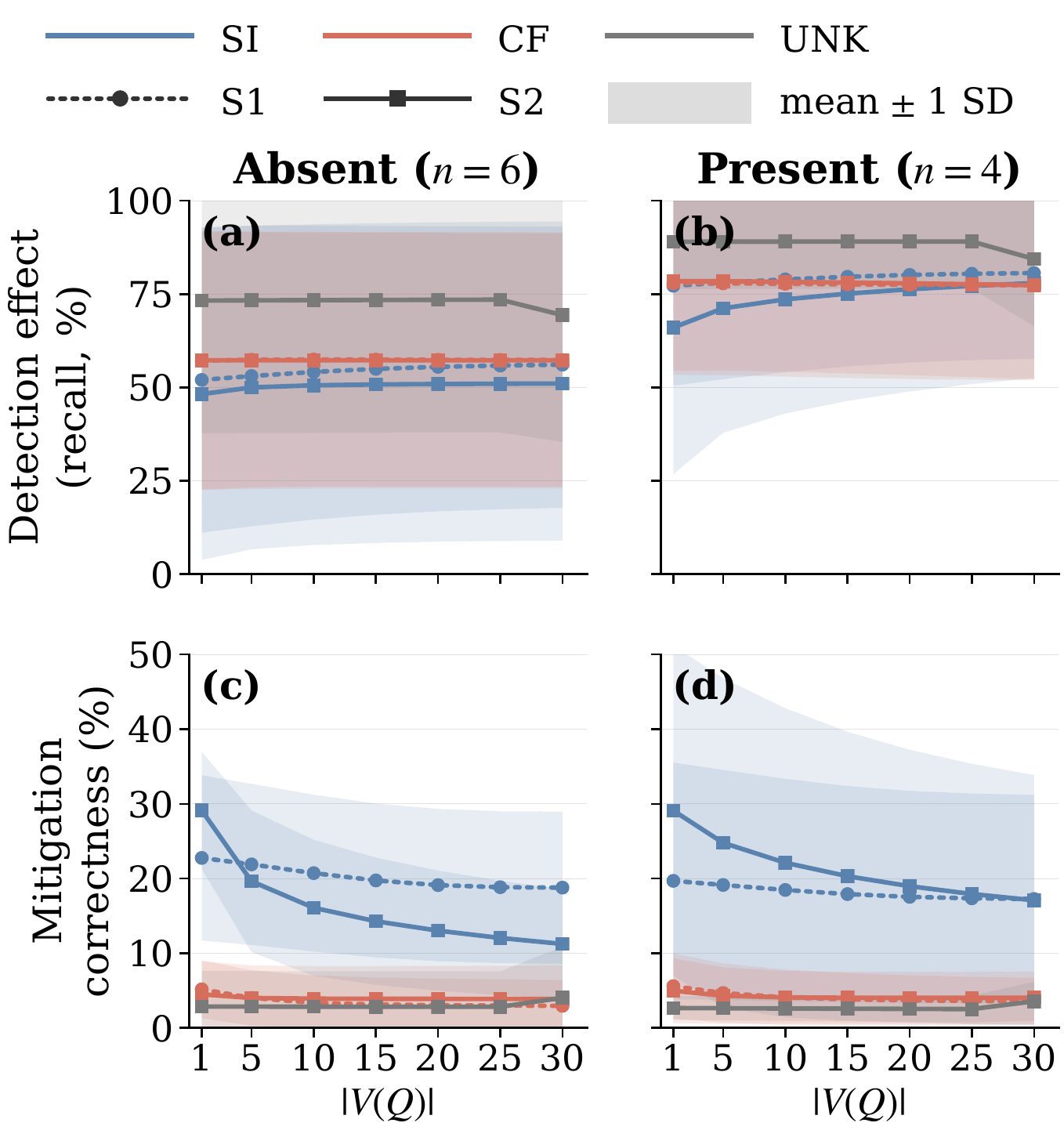}
        \caption{Knowledge Assumption}
        \label{fig:comparison_knowledge_assumption}
    \end{subfigure}
    \hfill
    \begin{subfigure}[t]{0.5\linewidth}
        \centering
        \vspace{1.5em}
        \includegraphics[width=\linewidth]{figures/comparison_intervention.pdf}
        \caption{Intervention Stage}
        \label{fig:comparison_intervention}
    \end{subfigure}
    \caption{\textbf{Comparison of existing methods.}}
    \label{fig:comparison}
\end{figure}

\ecoparagraph{Prompt sensitivity methods detect \CF{} almost on par with \SI{} under S1 but mitigate \SI{} only, performing the strongest \SI{} mitigation among all method dimensions (Fig.~\ref{fig:comparison_prompt_sensitivity}) .}
They detect \CF{} and \UNK{} well but mitigate them at near zero rate.
Notably, their \SI{} detection under S2 decreases from approximately $76\%$ to $67\%$ as $|\mathcal{V}(Q)|$ increases, the only notable decline in detection across the method dimensions. 
A possible explanation is that larger $|\mathcal{V}(Q)|$ filters out \SI{} errors readily exposed by rephrasing, leaving more difficult errors that prompt sensitivity methods struggle to detect.
Methods that do not model prompt sensitivity show the complementary pattern, with weaker detection but better \CF{} and \UNK{} mitigation. 
This suggests a potential composition in which prompt sensitivity identifies errors and stronger corrective methods that are outside the prompt levels mitigate the detected cases.

\ecoparagraph{Methods assuming knowledge is present (Present methods) outperform those assuming absent (Absent methods) in detection for all three types (\SI{}, \CF{}, and \UNK{}) and mitigation for hard \SI{} under S2 (Fig.~\ref{fig:comparison_knowledge_assumption}).}
Present methods also exhibit increasing \SI{} detection under S2 as $|\mathcal{V}(Q)|$ grows. 
This detection advantage is noteworthy because S2 labeling scheme is likely to remain difficult \SI{} cases after uncertainty elicitation.
Their \SI{} mitigation under S2 decreases yet remains at or above the corresponding S1 rate across $|\mathcal{V}(Q)|$. 
For Absent methods, \SI{} mitigation under S2 steadily decreases, even falling below the S1 rate at $|\mathcal{V}(Q)|{=}5$ and reaching roughly half of it at $|\mathcal{V}(Q)|{=}30$, despite relatively stable detection. 
This suggests that methods assuming accessible model knowledge remain effective at both detecting and mitigating these hard errors robustly.

\ecoparagraph{Only external knowledge mitigates \CF{} at a meaningful rate (Fig.~\ref{fig:comparison_intervention}).}
External knowledge methods achieve the most substantial \CF{} and \SI{} mitigation among the three stages, with only a limited decline as $|\mathcal{V}(Q)|$ increases. 
Post-hoc methods detect \UNK{} strongly but provide almost no mitigation correction for either \UNK{} or \CF{}, demonstrating that detecting an abstention does not itself correct an incorrect response. 
Inference-time intervention achieves the highest overall detection but comparatively weak mitigation, with S2 \SI{} mitigation declining even as \SI{} recall increases. The distinction is therefore pronounced. Inference-time methods are effective at detecting target errors without reliably mitigating them, whereas external knowledge methods are comparatively effective at mitigation despite more moderate detection.

\subsection{Uncertainty Estimation for Hallucinations}
\label{sec:uncertainty}

\definecolor{siorange}{RGB}{255,140,0}
\definecolor{cfpurple}{RGB}{138,43,226}
\definecolor{posgreen}{RGB}{34,139,34}

\begin{table}[h!]
\centering
\caption{
\textbf{Uncertainty estimation-based detection and mitigation performance}
(mean$\pm$std). 
}
\scriptsize
\setlength{\tabcolsep}{3pt}
\renewcommand{\arraystretch}{1.1}
\resizebox{\textwidth}{!}{%
\begin{tabular}{lccc ccc ccc ccc}
\toprule

& \multicolumn{6}{c}{\textbf{Detection}} 
& \multicolumn{6}{c}{\textbf{Mitigation}} \\

\cmidrule(lr){2-7}
\cmidrule(lr){8-13}

& \multicolumn{3}{c}{S1}
& \multicolumn{3}{c}{S2}
& \multicolumn{3}{c}{S1}
& \multicolumn{3}{c}{S2} \\

\cmidrule(lr){2-4}
\cmidrule(lr){5-7}
\cmidrule(lr){8-10}
\cmidrule(lr){11-13}

Metric
& CF & SI & UNK
& CF & SI & UNK
& CF & SI & UNK
& CF & SI & UNK \\

\midrule
Miss rate (\%) $\downarrow$
& 34.2$\pm$25.6
& 35.6$\pm$24.7
& \textbf{22.9$\pm$21.4}
& 50.4$\pm$33.5
& 35.9$\pm$24.9
& \textbf{30.8$\pm$23.0}
& --
& 27.9$\pm$37.6
& --
& 41.1$\pm$44.7
& \textbf{28.1$\pm$37.7}
& 29.6$\pm$38.1 \\

Hallucination score $\uparrow$
& 0.54$\pm$0.21
& 0.43$\pm$0.20
& \textbf{0.61$\pm$0.25}
& 0.42$\pm$0.29
& 0.42$\pm$0.21
& \textbf{0.57$\pm$0.19}
& -- & -- & --
& -- & -- & -- \\

Correctness (\%) $\uparrow$
& --
& --
& --
& --
& --
& --
& 3.9$\pm$7.3
& \textbf{29.5$\pm$7.5}
& 0.0$\pm$0.0
& 2.2$\pm$5.0
& \textbf{30.4$\pm$8.6}
& 0.4$\pm$0.7 \\

\bottomrule
\end{tabular}}
\label{tab:unk_combined}
\vspace{1em}
\end{table}

\begin{figure}[t!]
    \centering
    \begin{subfigure}[t]{0.7\linewidth}
        \centering
        \includegraphics[width=0.9\linewidth]{figures/comparison_uncertainty_representation.pdf}
        \caption{Uncertainty representation}
        \label{fig:comparison_uncertainty_representation}
    \end{subfigure}
    \hfill
    \begin{subfigure}[t]{0.4\linewidth}
        \centering
        \vspace{1.5em}
        \includegraphics[width=0.9\linewidth]{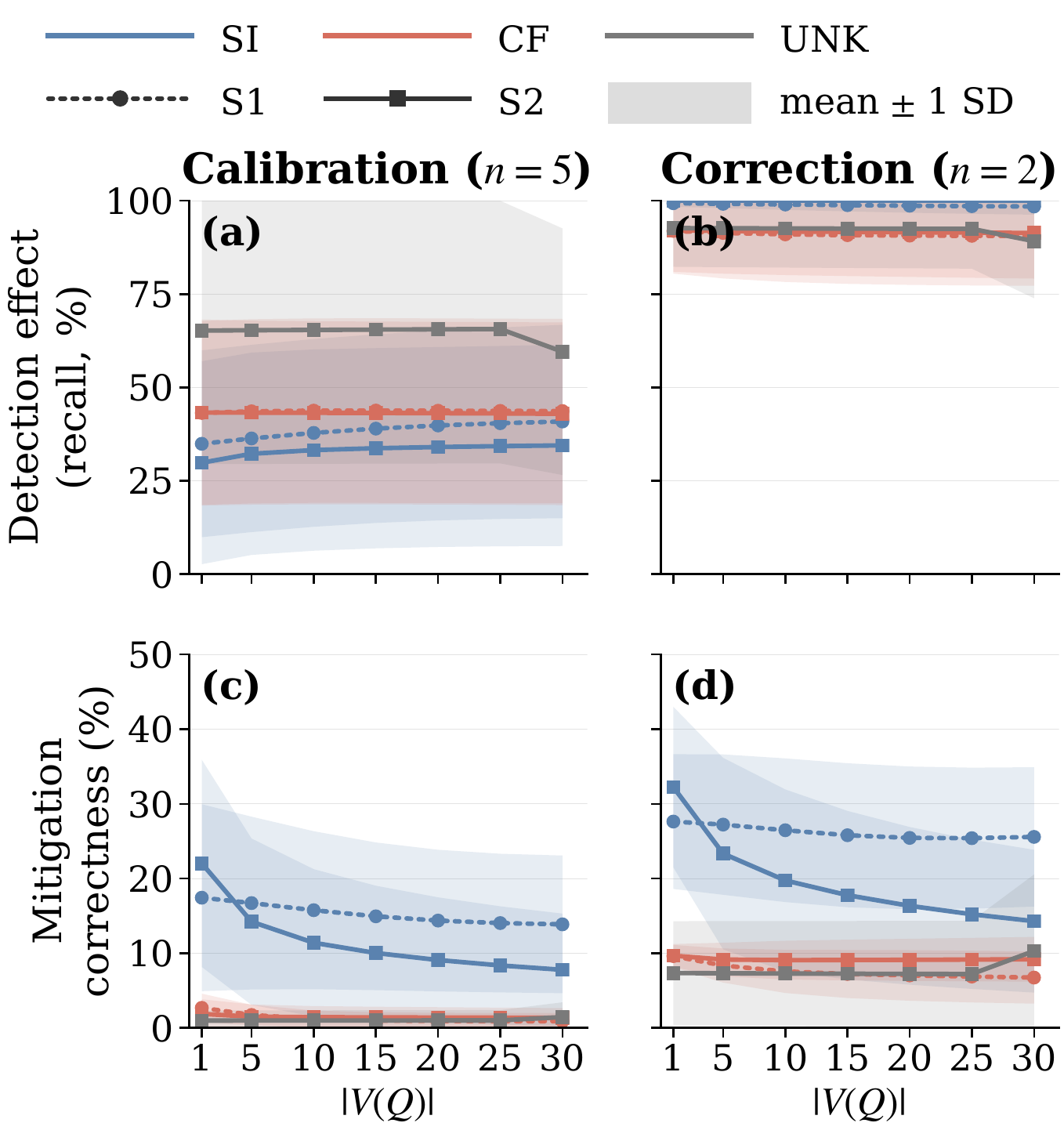}
        \caption{Uncertainty usage}
        \label{fig:comparison_uncertainty_usage}
    \end{subfigure}
    \caption{\textbf{Comparison of subset of existing methods that adopt uncertainty estimation.}}
    \vspace{-1em}
    \label{fig:comparison_uncertainty}
\end{figure}

\ecoparagraph{Uncertainty estimation detects \UNK{} best but mitigates only \SI{} (Table~\ref{tab:unk_combined}).}
Detection methods assign higher hallucination scores to \UNK{} than to \SI{} and \CF{}.
This means that the hallucination detection methods strongly align with the \UNK{} responses, consistently stronger than the \CF{} regardless of the uncertainty elicitation setting (i.e., both S1 and S2 labeling schemes). 
The gap arises as the uncertainty based methods rely on uncertainty cues, which \CF{} responses do not carry due to their confident incorrectness.

\ecoparagraph{Output-level signals are well-suited to \UNK{} while latent representations are promising for \CF{} (Fig.~\ref{fig:comparison_uncertainty_representation}).}
Verbal (output-level) uncertainty achieves the best detection on \UNK{} among uncertainty representations, while showing the worst detection on \CF{}.
Latent representations achieve the best detection on \CF{} among uncertainty representations, better than \SI{} across all $|\mathcal{V(Q)}|$.
The favorable patterns of latent methods hold in mitigation. 
Among the uncertainty representations, latent methods mitigate the most \CF{} and \UNK{} responses.
The advantage of using verbal representation does not translate to mitigation effect, however. 
Instead, semantic representation shows the best mitigation correctness on \SI{}.
Methods in the None group mitigate \CF{} even better than latent methods, but they rely on external knowledge or decoding instead of uncertainty.

\ecoparagraph{Correction methods mitigate \CF{} and \UNK{} slightly better than calibration methods (Fig.~\ref{fig:comparison_uncertainty_usage}).}
Mitigation methods that use uncertainty to correct the original response, achieve slightly higher mitigation effect on \CF{} and \UNK{} than Calibration methods, which use uncertainty primarily as a scoring signal. 
Mitigation methods therefore represent a promising step beyond calibration, yielding measurable gains in \CF{} and \UNK{} mitigation. However, these gains remain modest relative to their near ceiling detection of \SI{}, \CF{}, and \UNK{}, leaving substantial room to translate successful detection into correct answers.